\documentclass{article} 
\usepackage{iclr2025_conference,times}
\iclrfinalcopy

\fancypagestyle{arxivstyle}{
  \fancyhf{}
  \fancyfoot[C]{\thepage}
  \fancyheadoffset[L,R]{0.1in}
  \renewcommand{\headrulewidth}{0.4pt}
  \renewcommand{\footrulewidth}{0pt}
}

\renewcommand{\headrulewidth}{0.4pt}
\renewcommand{\footrulewidth}{0pt}

\usepackage{amsmath,amsfonts,bm}

\def\eqref#1{equation~\ref{#1}}

\def\1{\bm{1}}

\DeclareMathAlphabet{\mathsfit}{\encodingdefault}{\sfdefault}{m}{sl}
\SetMathAlphabet{\mathsfit}{bold}{\encodingdefault}{\sfdefault}{bx}{n}

\usepackage{float}
\usepackage{hyperref}
\usepackage{url}
\usepackage{booktabs}
\usepackage{placeins}
\usepackage{multirow}
\usepackage{booktabs,longtable,tabularx,array,listings,xcolor}
\usepackage{tcolorbox}
\usepackage[table]{xcolor}
\usepackage{graphicx}
\usepackage{titletoc}
\usepackage{wrapfig}
\usepackage{caption}
\usepackage{algorithm}
\usepackage{algpseudocode}

\definecolor{modelgray}{RGB}{205,209,215}
\definecolor{datagray}{RGB}{242,243,245}
\definecolor{nlpgblue}{RGB}{225,235,248}
\definecolor{qwenhead}{RGB}{226,218,244} 
\definecolor{qwenbody}{RGB}{247,245,252} 
\definecolor{qwenhl}{RGB}{237,231,250}   

\tcbuselibrary{skins}
\usepackage{amssymb}
\newcolumntype{Y}{>{\raggedright\arraybackslash}X}

\title{NLPG: Natural-Language Policy Gradients for Self-Evolving Language Agents}

\author{
Xu Liu\textsuperscript{1,†}
,
WenZhang Wei\textsuperscript{2,†}
,
Jun Cao\textsuperscript{2}
,
Dehua Peng\textsuperscript{2}
,
Huan Chen\textsuperscript{3}
,
Zhipeng Gui\textsuperscript{2,*}
,
Huayi Wu\textsuperscript{3}
\\
\\
\textsuperscript{1}School of Electronic Information, Wuhan University, Wuhan 430079, China
\\
\textsuperscript{2}School of Remote Sensing and Information Engineering, Wuhan University, 
\\
Wuhan 430079, China
\\
\textsuperscript{3}State Key Laboratory of Information Engineering in Surveying, 
\\
Mapping and Remote Sensing, Wuhan University, Wuhan 430079, China
\\
\\
\textsuperscript{†}Equal contribution
\quad
\textsuperscript{*}Corresponding author
\\
\texttt{\{xuliu1,zhipeng.gui\}@whu.edu.cn}
}
\begin{document}

\maketitle
\pagestyle{arxivstyle}
\thispagestyle{arxivstyle}
\vspace{-0.18in}

\begin{abstract}
Large language model agents increasingly rely on compound programs for retrieval, tool use, reasoning, and verification, yet their failures often arise from local procedural decisions. Existing reinforcement-learning and prompt-optimization approaches typically rely on scalar rewards or repeatedly modify entire prompts, making it difficult to capture and reuse procedural improvements while preserving a frozen agent. To address this problem, We propose \textbf{Natural-Language Policy Gradients (NLPG)}, an external policy-memory method for improving a fixed agent without changing its model parameters or program structure. NLPG diagnoses execution traces, propagates downstream feedback backward through the module graph, and converts recurring failures into route-local natural-language corrections that are aggregated into bounded policy updates for subsequent executions. Across six benchmarks covering memory, reasoning, instruction following, and evidence verification, NLPG also outperforms the strongest listed baseline for each benchmark by 8.71 percentage points on average. These results provide evidence that evaluated procedural experience can be transformed into local and interpretable policy updates, enabling continual improvement of frozen agents.
\end{abstract}
\section{INTRODUCTION}
\label{sec:introduction}

Large language models (LLMs) increasingly solve tasks as compound agents: a fixed program coordinates multiple model calls, retrieval operations, tools, and intermediate representations before producing an answer ~\cite{yao2023react,schick2023toolformer,wu2023autogen,khattab2024dspy}. This decomposition improves capability, but it also creates a credit-assignment problem. A failed final answer may result from an earlier query, an omitted entity, an incorrect memory write, or the final synthesis step. Updating the whole prompt after every failure therefore provides only coarse feedback and may modify parts of the agent that were not responsible for the error.

Parameter-based reinforcement learning optimizes scalar objectives but typically requires many rollouts and offers limited interpretability \cite{sutton2018reinforcement,shao2024deepseekmath}. Non-parametric prompt optimizers generate, refine, or evolve natural-language instructions based on task performance \cite{zhou2023ape,pryzant2023apo,yang2024opro,opsahlong2024miprov2,fernando2024promptbreeder,agrawal2026gepa}. Related adaptation methods exploit user corrections, self-feedback, verbal reflection, or cross-task experience \cite{madaan2022memprompt,madaan2023selfrefine,shinn2023reflexion,zhao2024expel}, while long-term memory systems primarily store and retrieve interaction content \cite{packer2023memgpt,park2023generative,chhikara2025mem0,fang2025lightmem,yang2026promem}. However, these approaches generally lack explicit credit assignment from an evaluated failure to the responsible module policy and its corresponding update in a compound agent.

We introduce \textbf{Natural-Language Policy Gradients (NLPG)}, a
parameter-free policy-evolution method for compound language agents. Its key
mechanism is graph-backward credit assignment: NLPG traces a failed outcome
through recorded execution dependencies and attributes it to the responsible
module or task-family policy rather than to the entire agent or a global
prompt. It keeps the benchmark program, model parameters, and retrieval
backend fixed, while injecting bounded procedural instructions into the
relevant invocation. After each evaluation pass, the diagnosed deviations and
execution dependencies generate candidate corrections, which are aggregated
across executions, filtered for safety and redundancy, and ranked according to their evidence support (cross-execution support for a correction), failure pressure (the normalized weight of the associated execution deviation), and critic confidence (the critic's confidence that the correction is reusable). Accepted corrections are activated only in a later round, enabling localized and interpretable policy evolution without changing the original agent
architecture.

\begin{wrapfigure}[16]{r}{0.56\textwidth}
    \vspace{-0.8\baselineskip}
    \centering
    \includegraphics[width=\linewidth]{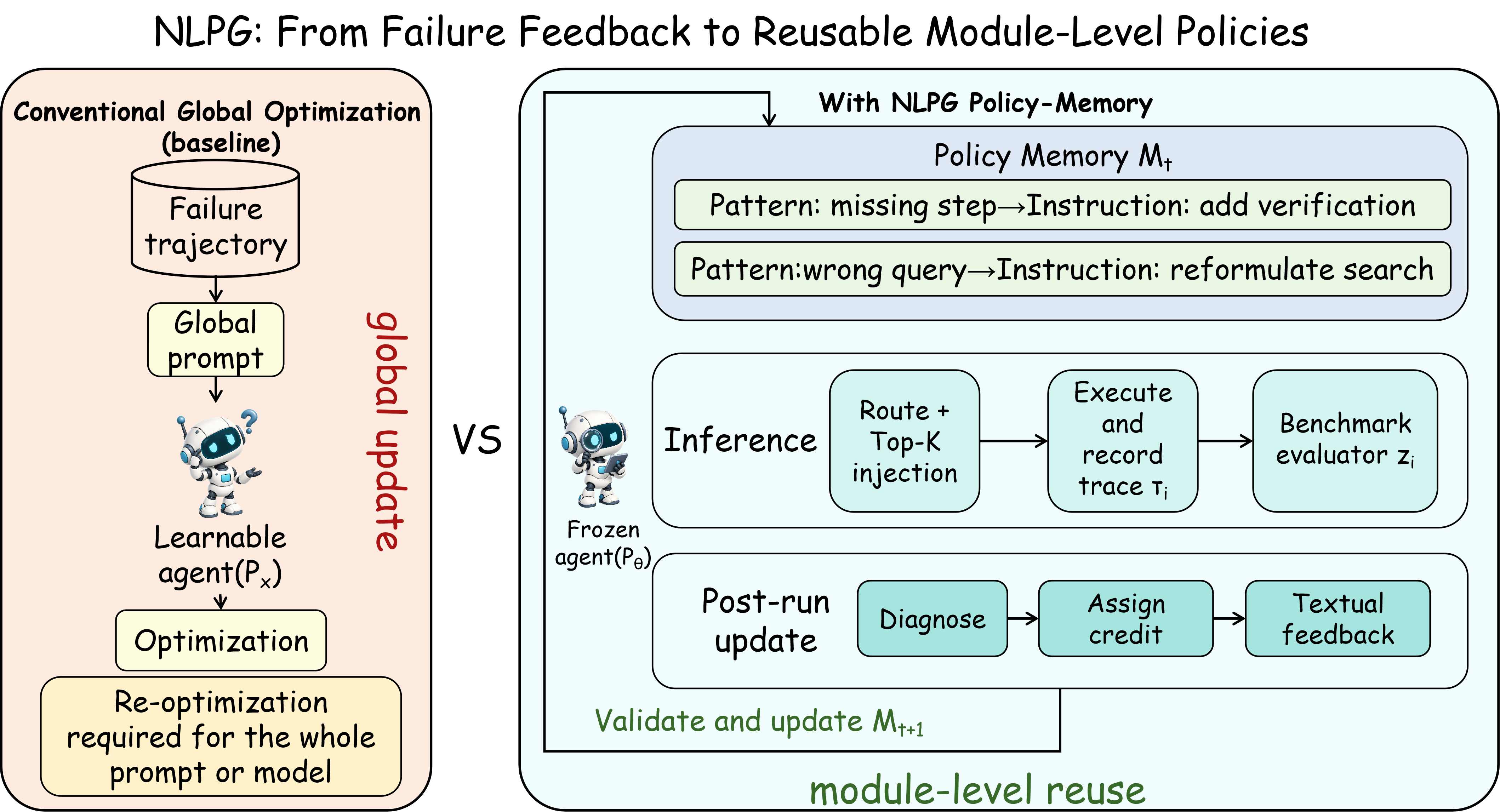}
    \caption{Conceptual comparison of global prompt-level updates and NLPG’s route-indexed policy updates.}
    \label{fig:introduction}
    \vspace{-0.8\baselineskip}
\end{wrapfigure}

NLPG separates policy execution from policy improvement in a compound agent. During execution, each module uses the policy associated with its route; after evaluation, the observed failure is traced back to the module policies that could have contributed to it, enabling module-local credit assignment. A language-based critic then converts this evidence into natural-language corrections, which are applied between evaluation passes. When execution dependencies are available, the same feedback can be propagated backward through the recorded module graph to produce more localized update directions.

We evaluate NLPG on six benchmarks spanning long-horizon memory and transfer:
LoCoMo, HaluMem, LongMemEval, HotpotQA, IFBench, and HoVer. Under
protocol-matched comparisons, NLPG improves the primary metric by 2.23--23.11
points over the strongest listed baseline, depending on the benchmark. In the
three-task transfer suite aligned with the GEPA evaluation setting
(HotpotQA, IFBench, and HoVer)~\cite{agrawal2026gepa}, NLPG averages 59.00 with Qwen3-8B, compared to 51.09 for GEPA (+7.91 percentage points) and averages 63.33 with GPT-4.1 Mini, compared to 59.43 for GEPA+Merge (+3.90 percentage points). Across these benchmarks, the improvements target procedural demands including multi-hop composition in HotpotQA, temporal grounding in LoCoMo, evidence routing in HoVer, and explicit output constraints in IFBench. This is a cross-benchmark pattern; the contribution of individual components is analyzed separately in the ablation experiments. Our contributions and key findings are summarized as follows:

\begin{itemize}
\item We introduce a parameter-free framework that evolves compound language
agents through external natural-language policy memory while keeping the
underlying program, model parameters, and retrieval system fixed.
\item We develop a natural-language gradient mechanism that attributes
diagnosed execution deviations along recorded dependencies and converts them
into reusable, route-local policy corrections.
\item We conduct a protocol-matched evaluation across six benchmarks and show
consistent gains in long-horizon memory use, multi-hop reasoning, instruction
following, and evidence-grounded verification under different model settings.
\end{itemize}

\section{Preliminaries}
\label{sec:preliminaries}

This section introduces the execution process of the self-evolving agent and defines the notation used in the remainder of the method. A discussion of related work is provided in Appendix~\ref{sec:related_work}.

\paragraph{Self-evolving compound agent.}
Let $P_{\theta}$ denote a self-evolving compound agent, where $\theta$ denotes the underlying model parameters and the fixed program configuration. Given a task input $x_i$ from a particular task category, the agent consults the current policy memory $M_t$ and invokes the required program modules according to its fixed execution procedure. The agent produces a task response $y_i$ and an evaluation result $z_i$ that records whether the execution satisfies the task requirements and, when available, the diagnosed failure information. The observable execution trace is denoted by $\tau_i$ (Sutton and Barto, 2018; Yao et al., 2023; Liu et al., 2024). After the current evaluation round is completed, the agent extracts effective policy instructions from the collected traces and uses them to construct the next memory version $M_{t+1}$. The model parameters and the fixed program configuration remain unchanged throughout this process.

We use the following notation:

\begin{itemize}
    \item \textbf{Program module.} A program module $b$ is a callable unit that performs one explicit operation in the agent's execution procedure, such as retrieval, external tool use, verification, or result parsing. For the $j$-th invocation in example $i$, the module produces an explicit output $o_{i,j}$.

    \item \textbf{Policy memory.} The policy memory is indexed by a task category or a program module. We write $M_t=\{M_t^a\}_{a\in\mathcal{A}}$, where $M_t^a$ contains short natural-language instructions describing how the corresponding category or module should execute its operation. These instructions may specify, for example, how to construct a retrieval query, organize intermediate evidence, or formulate the final answer. Policy memory stores procedural experience rather than task-specific reference answers.

    \item \textbf{Execution trace.} For task $i$, the observable execution trace records the task input, the explicit outputs produced during execution, and the final response:
    \begin{equation}
        \tau_i=\left(x_i,\{o_{i,j}\}_{j=1}^{J_i},y_i\right),
        \label{eq:prelim_trace}
    \end{equation}
    where $J_i$ is the number of module invocations. The trace may contain generated queries, retrieved evidence, tool results, and intermediate module outputs, but does not include hidden model reasoning.

    \item \textbf{Failure.} A failure is an identifiable deviation between the execution or final response and the task requirements that can be attributed to one or more relevant modules. Examples include omitting a key entity from a query, failing to retrieve a necessary document, or producing an answer that is unsupported by the available evidence. A failure is represented by a failure signature $s$ and may be associated with a set of policy-memory indices.

    \item \textbf{Benchmark evaluation.} For benchmark $\mathcal{B}$, an evaluator uses the execution trace and the benchmark's reference information, such as annotated evidence, task constraints, a gold answer, or scoring rules, to produce $z_i$. The reference information is used to evaluate the execution and diagnose failures; it is not stored as policy memory.
\end{itemize}

The self-evolution problem studied in this paper keeps $P_{\theta}$ and its execution procedure fixed, while updating the external policy memory. Given evaluated executions $(\tau_i,z_i)$, the goal is to convert recurring failures into natural-language update directions associated with the relevant memory indices and to retain only updates supported by multiple examples. Formally, the next policy-memory version is constructed as
\begin{equation}
    M_{t+1}=\operatorname{Update}\!\left(M_t,\{(\tau_i,z_i)\}_{i=1}^{N}\right).
    \label{eq:prelim_memory_update}
\end{equation}
where $M_t$ is fixed during the evaluation round and $M_{t+1}$ is loaded only in a subsequent round. The main challenge is therefore to determine both \textbf{where} a failure should update the policy memory and \textbf{how} the associated instruction should be changed. NLPG addresses this challenge by propagating failure information along the observed module dependencies and aggregating the resulting natural-language update directions across examples.

\begin{figure}[t]
    \centering
    \includegraphics[width=\linewidth]{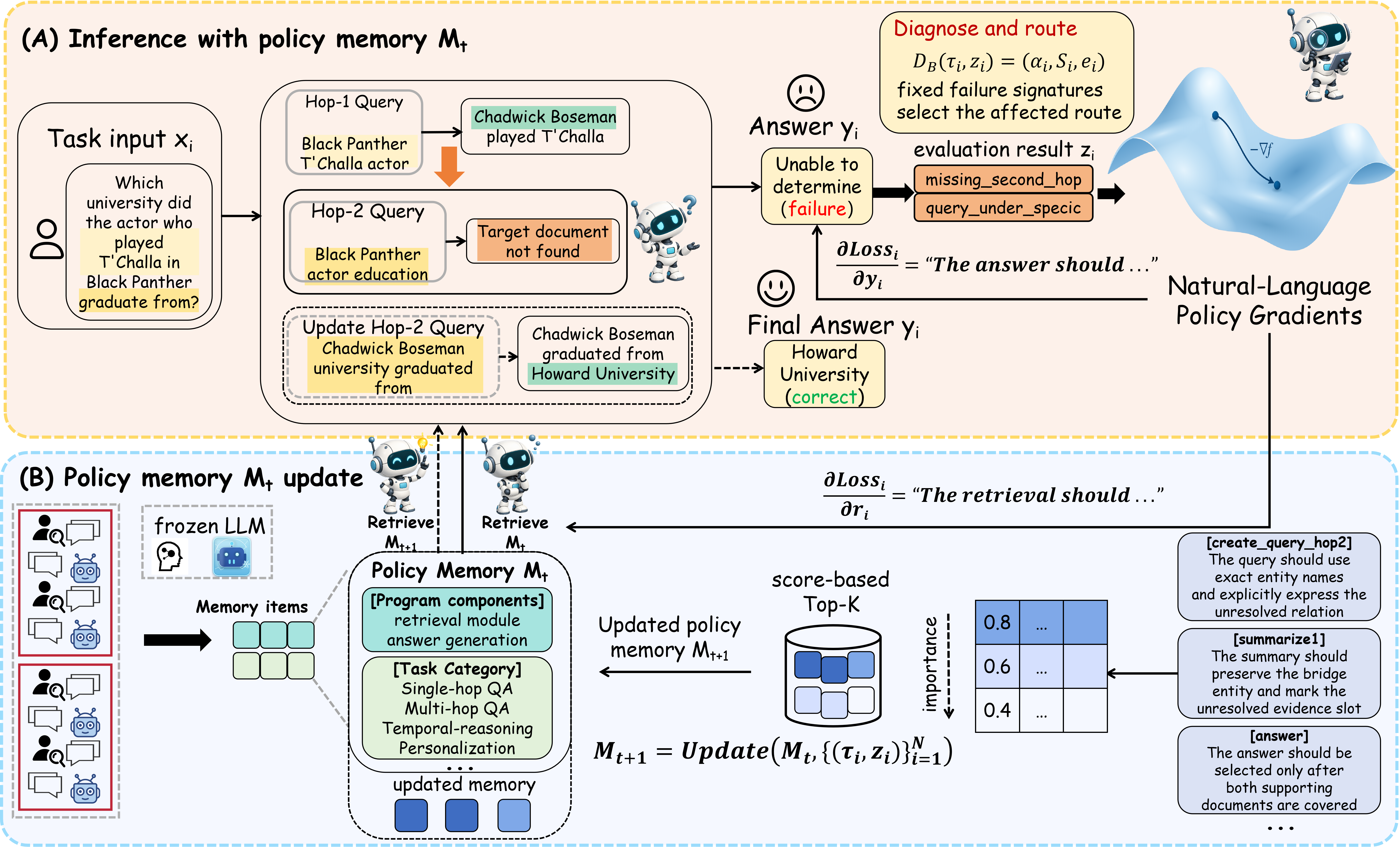}
    \caption{\textbf{Overview of NLPG.}
    \textbf{(A) Inference and failure diagnosis.} The compound agent retrieves module-local policies from the current policy memory $M_t$ and injects them into the corresponding module calls. 
    \textbf{(B) Policy-memory update.} NLPG aggregates routed language gradients
    across evaluated trajectories, scores and selects reusable instructions,
    and stores them under the corresponding program modules and task categories
    to construct $M_{t+1}$. }
    \label{fig:method}
\end{figure}
\section{Method}
\label{sec:method}

As shown in Figure~\ref{fig:method}, NLPG updates the procedural policies of a fixed compound agent while keeping its model, program, and retrieval index unchanged. Section~\ref{sec:nlpg_policy_memory} introduces routed policy memory, Section~\ref{sec:nlpg_textual_gradients} presents natural-language policy gradients, and Section~\ref{sec:nlpg_aggregation} describes cross-example policy aggregation and validation.

\subsection{NLPG Policy Memory}
\label{sec:nlpg_policy_memory}
To improve a compound agent without retraining its underlying models, NLPG treats procedural experience as an external, updateable source of improvement. Let \(P_{\theta}\) denote the compound agent, where \(\theta\) includes the
model parameters and program configuration that remain fixed during
self-evolution. Throughout this paper, a \textbf{task instance} denotes one
benchmark item supplied to the agent, while an \textbf{execution} denotes the
complete processing of that instance, including all module invocations, the
final response, and its evaluation. At policy version \(t\), the agent receives
task instance \(x_i\), consults the external policy memory \(M_t\), and executes
the fixed program.

\paragraph{Policy routing and candidate construction.}
For module-routed tasks, $a_{i,j}$ denotes the module invoked at step $j$ of example $i$; for task-family-routed tasks, it denotes the example's task family. The selected block $M_t^{a_{i,j}}$ contains reusable instructions for that route. When enabled, a shared block $M_t^{\mathrm{gen}}$ supplements it with route-independent rules on evidence sufficiency, output format, and
unsupported conclusions.

For the current invocation, NLPG reads the route-specific policy memory $M_t^{a_{i,j}}$ and, when available, combines it with the shared policy memory $M_t^{\mathrm{gen}}$. We set $M_t^{\mathrm{gen}}=\varnothing$ when a benchmark
does not define shared policy memory. After removing instructions with identical normalized text, the candidate policy set is
\begin{equation}
\mathcal{C}_{t,i,j}
=
\operatorname{Dedup}
\!\left(
M_t^{a_{i,j}}\cup M_t^{\mathrm{gen}}
\right),
\label{eq:nlpg_candidate_set}
\end{equation}
where \(\mathcal{C}_{t,i,j}\) is the complete candidate pool for the current
invocation. It contains the instructions attached to route \(a_{i,j}\) and,
when defined, the instructions shared across routes; an instruction appearing
in both memories is retained only once after text normalization. The pool is
used for selection rather than inserted in full: candidates are ranked by
their stored utility and tie-breaking scores, and only the highest-ranked
\(K\) instructions are formatted and added to the module context.

\paragraph{Policy selection.}
Each candidate $m\in\mathcal{C}_{t,i,j}$ contains an instruction and a historical priority score $w_t(m)$ derived from previous evaluated executions. NLPG ranks the candidates by this score and selects at most $K$ instructions:
\begin{equation}
\mathcal{P}_{t,i,j}
=
\operatorname{TopK}_{w_t}
\!\left(\mathcal{C}_{t,i,j},K\right).
\label{eq:nlpg_topk}
\end{equation}

The selected set determines which policy instructions are injected into the
current module invocation. The benchmark-specific values of \(K\) and the
policy-selection configurations used for the reported runs are provided in
Appendix~\ref{app:implementation}.

\paragraph{Policy injection and module execution.}
Let $c_{i,j}^{\mathrm{base}}$ denote the original execution context constructed by the task program without NLPG policies. It contains the module instruction, the current task input, and any outputs passed from preceding invocations. The selected instructions $\mathcal{P}_{t,i,j}$ are retrieved from policy memory, formatted as a bounded policy addendum, and appended to the original context:
\begin{equation}
c_{i,j}(M_t)
=
c_{i,j}^{\mathrm{base}}
\mathbin{\Vert}
\operatorname{Format}\!\left(\mathcal{P}_{t,i,j}\right),
\label{eq:nlpg_inference_context}
\end{equation}
where $\Vert$ denotes text concatenation. The module processes the resulting
context and produces output $o_{i,j}$, which may subsequently be consumed by
downstream modules.

After $J_i$ invocations, the agent returns the final response $y_i$. The
observable execution trace is
\begin{equation}
\tau_i=
\left(
x_i,
\{a_{i,j},c_{i,j},o_{i,j}\}_{j=1}^{J_i},
y_i
\right),
\label{eq:nlpg_trace}
\end{equation}
where each invocation records its policy route, execution context, and explicit
output. The trace may contain generated queries, retrieved evidence, tool
outputs, and other observable intermediate results. The benchmark evaluator applies the evaluation to the completed execution and returns the evaluation result $z_i$.

\subsection{Natural-Language Policy Gradients}
\label{sec:nlpg_textual_gradients}

Task-level evaluation reveals whether an execution succeeds, but not which
module caused a failure. NLPG therefore combines the evaluation result with the observed module dependencies to identify responsible modules and generate route-specific corrections. We call this natural-language description of a local error and its correction direction a \textbf{natural-language policy gradient}.

\paragraph{Failure diagnosis.}
We define an evaluated execution as the complete process of running one task example, from its input and module invocations to its final output and evaluation. For benchmark $\mathcal B$, the diagnosis function maps execution trace $\tau_i$ and evaluation result $z_i$ to
\begin{equation}
\mathcal D_{\mathcal B}(\tau_i,z_i)
=
(\alpha_i,\mathcal S_i,e_i),
\label{eq:nlpg_diagnosis}
\end{equation}
where \(\alpha_i\) is the failure weight, \(\mathcal{S}_i\) is the set of
diagnosed execution-deviation types, and \(e_i\) is the identifier of the
evaluated execution. This identifier is subsequently used to count how many
distinct executions provide evidence for the same candidate correction. In our
experiments, \(\alpha_i=0\) for correct executions and \(\alpha_i=1\) for
failed executions; records without a valid evaluation are excluded from policy
construction.

\paragraph{Backward propagation of natural-language policy gradients.}

Inspired by TextGrad~\cite{yuksekgonul2025textgrad}, NLPG propagates
task-level language feedback \(\mathcal{L}_i\) through the observed
module-dependency graph in reverse topological order. At a terminal module,
the downstream feedback is initialized with the task-level evaluation. At an
intermediate module, \(F_{i,j}^{\mathrm{down}}\) collects the feedback messages
returned by its immediate successor modules. The critic examines these
messages together with the current module's input, output, and execution
context, produces a localized textual correction, and forwards feedback only
to the module's recorded predecessors. This operation propagates feedback
within one execution graph; the grouping of corrections across distinct
executions is performed separately in
Section~\ref{sec:nlpg_aggregation}. Consequently, for invocation output
\(o_{i,j}\), the language-level gradient is defined as
\begin{equation}
\frac{\partial^{\mathrm{text}}\mathcal L_i}
     {\partial y_{i,j}}
\mathrel{\stackrel{\mathrm{def}}{=}}
\nabla_{\mathrm{LLM}}
\!\left(
o_{i,j},
F_{i,j}^{\mathrm{down}}
\right),
\label{eq:nlpg_output_gradient}
\end{equation}
where $\partial^{\mathrm{text}}$ denotes a language-level feedback transformation rather than a numerical derivative, and $\nabla_{\mathrm{LLM}}$ denotes the critic operation that transforms the
current output and downstream feedback into a textual description of how the response $y_{i,j}$ should be improved. When an execution graph is available, NLPG propagates this feedback from
terminal modules backward in reverse topological order. At each module, the
critic uses the module's output and accumulated downstream feedback to
formulate a local correction and pass relevant feedback to its immediate
predecessors. Propagation terminates along a branch when no further feedback
is assigned, so corrections follow dependencies observed in the execution.

Conditioned on the output-level feedback above, NLPG attributes the required change to the routed policy that produced the module output. We use the
dependency\(M_t^{a_{i,j}} \rightarrow o_{i,j} \rightarrow \mathcal{L}_i\) as a structural description of this attribution. The corresponding route-level natural-language gradient is defined as
\begin{equation}
g_{i,j}^{\mathrm{text}}
\equiv
\frac{\partial^{\mathrm{text}}\mathcal L_i}
     {\partial r_{i,j}}
\mathrel{\stackrel{\mathrm{def}}{=}}
\nabla_{\mathrm{LLM}}
\!\left(
M_t^{a_{i,j}},
c_{i,j},
o_{i,j},
F_{i,j}^{\mathrm{down}}
\right),
\label{eq:nlpg_policy_gradient}
\end{equation}

Here, $r_{i,j}$ denotes the policy instruction retrieved for the $j$-th
invocation, while $M_t^{a_{i,j}}$ denotes the route-specific policy memory
from which this instruction is selected; $c_{i,j}$ and $o_{i,j}$ denote the invocation context and the resulting output, respectively, while $F_{i,j}^{\mathrm{down}}$ carries task-level or downstream feedback. The operator $\partial^{\mathrm{text}}$ denotes a language-level transformation rather than a numerical derivative. The critic routes each correction along the observed dependencies, after which NLPG filters and aggregates equivalent instructions across executions. Their priority reflects support, failure weight, and critic confidence, while their effectiveness is assessed by subsequent validation.

\subsection{Cross-Execution Aggregation and Policy Update}
\label{sec:nlpg_aggregation}

A correction derived from a single execution may reflect an incidental property of the example or its execution path. NLPG therefore aggregates accepted natural-language policy gradients across executions before updating policy memory. Gradients are grouped by route $a$, failure type $s$, and normalized correction text $u$. For each group, let $g_{i,j}^{\mathrm{text}}$ denote the representative gradient, selected by descending failure weight and then critic confidence, where $i,j$ identify the execution and invocation that produced it.

Let \(\mathcal{E}_{a,s,u}\) denote the set of distinct evaluated-execution
identifiers associated with correction \(u\), and let $\mathcal E_a$ denote the set of distinct evidence identifiers associated with all accepted corrections on route $a$. Let $N_a$ be the number of valid evaluated records attributed to route $a$. NLPG assigns the candidate correction the score
\begin{equation}
w_{a,s,u}
=
\frac{
|\mathcal E_{a,s,u}|
}{
\max\!\left(N_a,|\mathcal E_a|,1\right)
}
\,\alpha_i\,\kappa_{i,j}.
\label{eq:nlpg_candidate_score}
\end{equation}
The three factors represent cross-execution support, failure weight, and critic confidence. We set $\alpha_i=1$ for failures and $0$ otherwise, so the score ranks corrections for retention. The optimizer converts each grouped correction and its diagnosis into a concise route-specific instruction. After safety filtering, accepted instructions are merged with the current
policy memory to form the candidate successor \(\widehat{M}_{t+1}\).. NLPG promotes this candidate only if paired held-out validation satisfies the gain and regression constraints; otherwise, it retains $M_t$. The promoted memory provides the Top-$K$ policies in the next round.

Overall, NLPG turns execution feedback into validated, route-specific policy
updates while keeping the agent's model and program fixed. By making each
correction explicit and reusable, it supports interpretable improvement
across successive rounds of agent execution.
\section{Experiments}
\label{sec:experiments}

\subsection{Experimental Setup}
\label{sec:experimental_setup}

\paragraph{Experimental Details.}
We follow the released protocol for each benchmark. Within every comparison,
all methods use matched examples, models, prompts, retrieval settings, and
scorers; published results are reused only when these conditions agree and are
otherwise recomputed in the NLPG evaluation harness. The policy version remains
fixed within each evaluation pass. Full configurations are provided in
Appendix~\ref{app:experimental-setup}.

\paragraph{Datasets \& Baseline Methods.}
We evaluate memory on LoCoMo~\cite{maharana2024locomo}, HaluMem
~\cite{chen2025halumem}, and LongMemEval~\cite{wu2025longmemeval}, against the
memory-agent baselines reported for their aligned protocols, including
Mem0~\cite{chhikara2025mem0},MemR$^3$~\cite{du2025memr3}, LightMem~\cite{fang2025lightmem}, and
ProMem~\cite{yang2026promem}. Transfer is evaluated on the GEPA-aligned
HotpotQA~\cite{yang2018hotpotqa}, IFBench~\cite{pyatkin2025ifbench}, and
HoVer~\cite{jiang2020hover} suite against the task baseline, GRPO,
MIPROv2~\cite{opsahlong2024miprov2}, TextGrad~\cite{yuksekgonul2025textgrad},
Trace/OptoPrime~\cite{cheng2024trace}, and GEPA~\cite{agrawal2026gepa}. The
complete baseline lists and data construction are deferred to
Appendix~\ref{app:experimental-setup}.

\paragraph{Metrics.}
LoCoMo and LongMemEval use QA accuracy; HaluMem uses Memory Integrity, Memory
Accuracy, and QA Accuracy. HotpotQA uses normalized exact match, IFBench the
official strict checker, and HoVer its evidence-grounded verification score.
All values are percentages, and transfer-suite averages are arithmetic means
over the three tasks. Auxiliary metrics and exact scoring rules are reported
in Appendix~\ref{app:experimental-setup}. Complete results for HaluMem and LongMemEval are reported in Appendices~\ref{app:halumem-details} and~\ref{app:longmemeval-details},
respectively.

\subsection{Long-Horizon Conversational Memory}
\label{sec:locomo_results}

\begin{table}[H]
\centering
\scriptsize
\setlength{\tabcolsep}{3.5pt}
\renewcommand{\arraystretch}{1.08}
\caption{LoCoMo results under the aligned evaluation protocol. Parenthesized values show improvements over the strongest baseline in each column.}
\label{tab:locomo_results}
\begin{tabular}{c|l|ccccc}
\toprule
\textbf{LLM} & \textbf{Method}
& \textbf{1. Multi-Hop}
& \textbf{2. Temporal}
& \textbf{3. Open-Domain}
& \textbf{4. Single-Hop}
& \textbf{Overall} \\
\midrule
\multirow{10}{*}{\rotatebox[origin=c]{90}{GPT-4o-mini}}
& MemoryOS
& 56.50 & 37.18 & 40.28 & 62.43 & 54.70 \\

& Mem0
& 58.16 & 55.45 & 40.62 & 66.71 & 61.00 \\

& MemU
& 62.41 & 33.96 & 46.88 & 72.77 & 61.15 \\

& MemOS
& 69.15 & 72.27 & 60.42 & 81.45 & 75.87 \\

& HiMem
& 70.92 & 74.77 & 54.86 & 89.22 & 80.71 \\

& Zep
& \textbf{71.99} & 74.45 & \textbf{66.67} & 88.11 & 81.06 \\

& TiMem
& 62.20 & \textbf{77.63} & 52.08 & 81.43 & 75.30 \\

& TSM
& 66.67 & 71.03 & 58.33 & 84.30 & 76.69 \\

& MemR$^3$
& 71.39 & 76.22 & 61.11 & \textbf{89.44} & \textbf{81.55} \\

\cmidrule(lr){2-7}

& \textbf{NLPG (ours)}
& \textbf{75.53}\,\textcolor{red}{(+4.14)}
& \textbf{86.92}\,\textcolor{red}{(+10.70)}
& \textbf{70.83}\,\textcolor{red}{(+9.72)}
& \textbf{92.29}\,\textcolor{red}{(+2.85)}
& \textbf{86.72}\,\textcolor{red}{(+5.17)} \\

\bottomrule
\end{tabular}
\end{table}
The category-level results indicate that NLPG is most useful when answering
requires a sequence of memory operations rather than a single semantic match.
The gains on multi-hop and temporal questions show that route-local policies
help the agent combine evidence and preserve time-dependent distinctions across
retrieval and answer synthesis. The smaller improvement on single-hop
questions is consistent with their lower procedural complexity: these questions
often require retrieving one directly relevant fact, leaving less room for
module coordination to improve the result.

\subsection{Transfer to Reasoning, Instruction Following, and Verification}
\label{sec:gepa_aligned_results}

We next test transfer beyond conversational memory on the GEPA-aligned
HotpotQA/IFBench/HoVer suite. The comparison includes MIPROv2~\cite{opsahlong2024miprov2},
TextGrad~\cite{yuksekgonul2025textgrad}, Trace/OptoPrime~\cite{cheng2024trace},
and GEPA~\cite{agrawal2026gepa}. Results are reported for Qwen3-8B and
GPT-4.1 Mini.

\begin{table}[H]
\centering
\scriptsize
\caption{Cross-model generalization under the GEPA evaluation suite~\cite{agrawal2026gepa}. NLPG is evaluated with Qwen3-8B~\cite{yang2025qwen3} and GPT-4.1 Mini~\cite{openai2025gpt41} on HotpotQA, IFBench, and HoVer.}
\label{tab:gepa_model_generalization}
\setlength{\tabcolsep}{15pt}
\begin{tabular}{lccccc}
\toprule
\rowcolor{qwenhead}
\multicolumn{6}{c}{
  \textbf{\raisebox{-0.15em}{\includegraphics[height=1.1em]{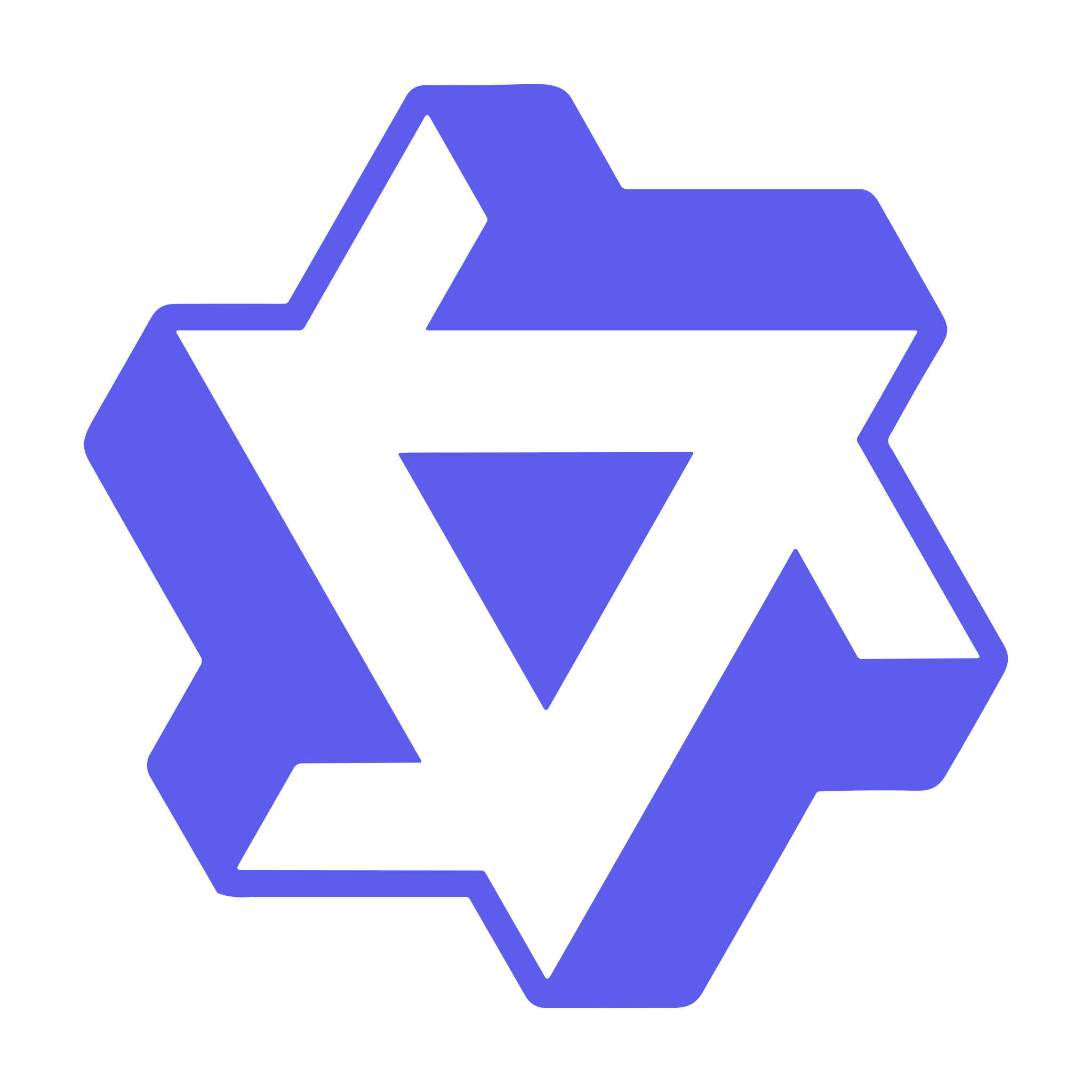}}\,Qwen3-8B}
} \\
\midrule
\rowcolor{qwenbody}
\textbf{Method} & \textbf{HotpotQA} & \textbf{IFBench} & \textbf{HoVer} & \textbf{3-task Avg.} & \textbf{Improvement} \\
\midrule
\rowcolor{qwenbody}
Baseline          & 42.33 & 36.90 & 35.33 & 38.19 & -- \\
\rowcolor{qwenbody}
GRPO              & 43.33 & 35.88 & 38.67 & 39.29 & +1.11 \\
\rowcolor{qwenbody}
MIPROv2           & 55.33 & 36.22 & 47.33 & 46.29 & +8.11 \\
\rowcolor{qwenbody}
GEPA              & 62.33 & 38.61 & 52.33 & 51.09 & +12.90 \\
\rowcolor{qwenbody}
GEPA+Merge        & 64.33 & 28.23 & 51.67 & 48.08 & +9.89 \\
\rowcolor{qwenhl}
NLPG              & \textbf{73.67} & \textbf{44.00} & \textbf{59.33} & \textbf{59.00} & \textbf{+20.81} \\
\midrule
\rowcolor{modelgray}
\multicolumn{6}{c}{
  \textbf{\raisebox{-0.15em}{\includegraphics[height=1.1em]{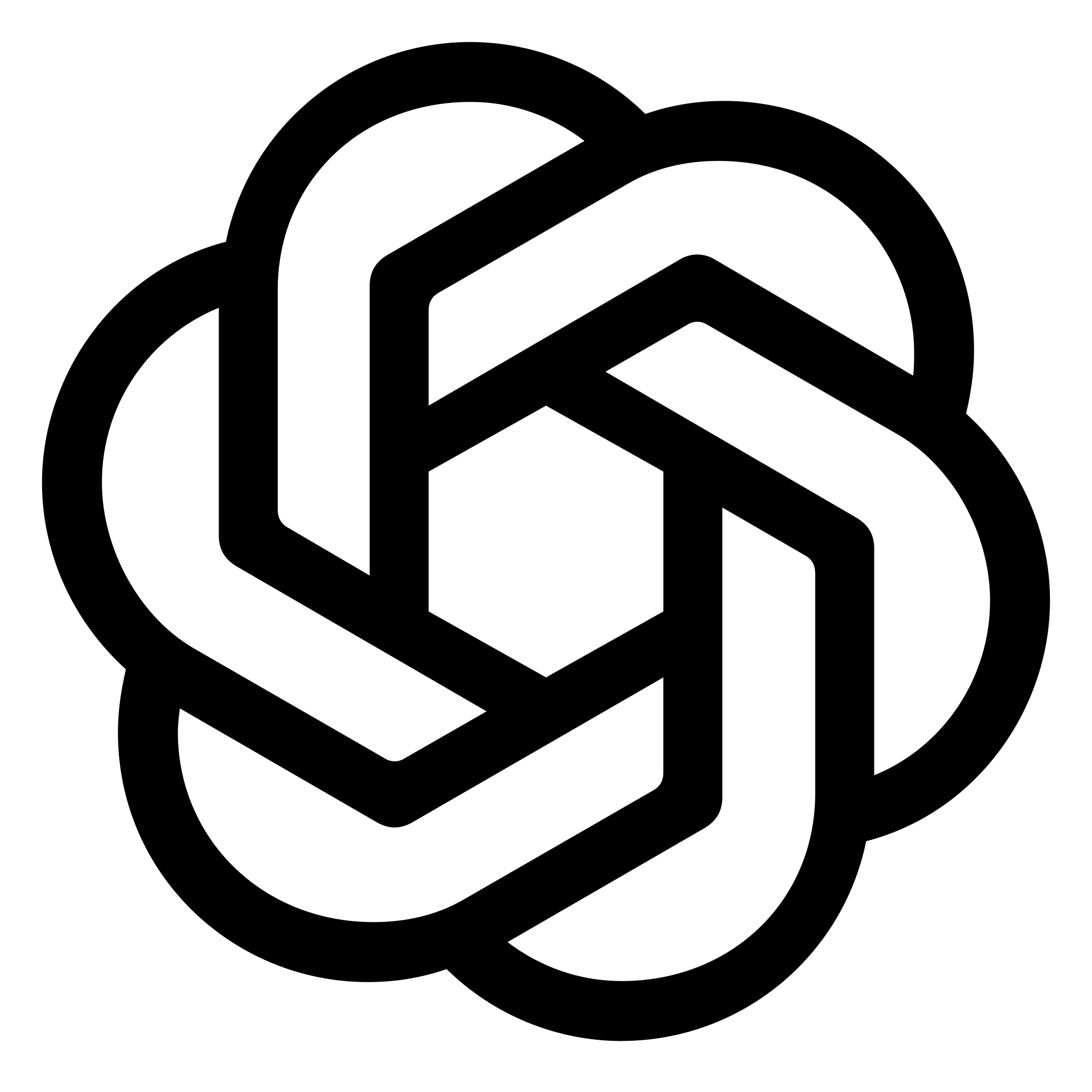}}\,GPT-4.1 Mini}
} \\
\midrule
\rowcolor{datagray}
\textbf{Method} & \textbf{HotpotQA} & \textbf{IFBench} & \textbf{HoVer} & \textbf{3-task Avg.} & \textbf{Improvement} \\
\midrule
\rowcolor{datagray}
Baseline          & 38.00 & 47.79 & 46.33 & 44.04 & -- \\
\rowcolor{datagray}
Trace (OptoPrime) & 60.33 & 51.19 & 46.00 & 52.51 & +8.47 \\
\rowcolor{datagray}
MIPROv2-No-Demos  & 38.00 & 52.04 & 51.33 & 47.12 & +3.08 \\
\rowcolor{datagray}
MIPROv2           & 58.00 & 49.15 & 48.33 & 51.83 & +7.79 \\
\rowcolor{datagray}
TextGrad          & 62.33 & 48.64 & 47.67 & 52.88 & +8.84 \\
\rowcolor{datagray}
GEPA              & 69.00 & 52.72 & 51.67 & 57.80 & +13.76 \\
\rowcolor{datagray}
GEPA+Merge        & 65.67 & 55.95 & 56.67 & 59.43 & +15.39 \\
\rowcolor{datagray}
GEPA-Qwen-Opt     & 65.67 & 49.83 & 54.67 & 56.72 & +12.68 \\
\rowcolor{nlpgblue}
NLPG              & \textbf{72.67} & \textbf{58.00} & \textbf{59.33} & \textbf{63.33} & \textbf{+19.29} \\
\bottomrule
\end{tabular}
\end{table}

NLPG is best on all three tasks for both backbones, averaging 59.00 with
Qwen3-8B and 63.33 with GPT-4.1 Mini. Diagnostic breakdowns are provided in
Appendix~\ref{app:ifbench-hover-details}. The transfer results show that the benefit is not tied to conversational-memory retrieval. NLPG improves all three tasks for both backbones, including instruction following, multi-hop retrieval, and evidence-grounded verification. Because the baseline and NLPG use the same backbone, examples, task procedure, retrieval settings, generation limits, and evaluation metric in each comparison, NLPG does not receive extra model calls or a larger inference budget. Under the matched comparison protocol, the results are consistent with a contribution from the external policy memory and update mechanism, rather than from a different answer format or more computation.

\subsection{Cross-Benchmark Analysis}
\label{sec:cross_benchmark_analysis}

Across the six independently held-out evaluations, NLPG improves both
long-horizon memory use and compound-task execution. Policies are constructed
and selected without access to final-test examples.

\begin{figure}[t]
    \centering
    \includegraphics[width=\linewidth]{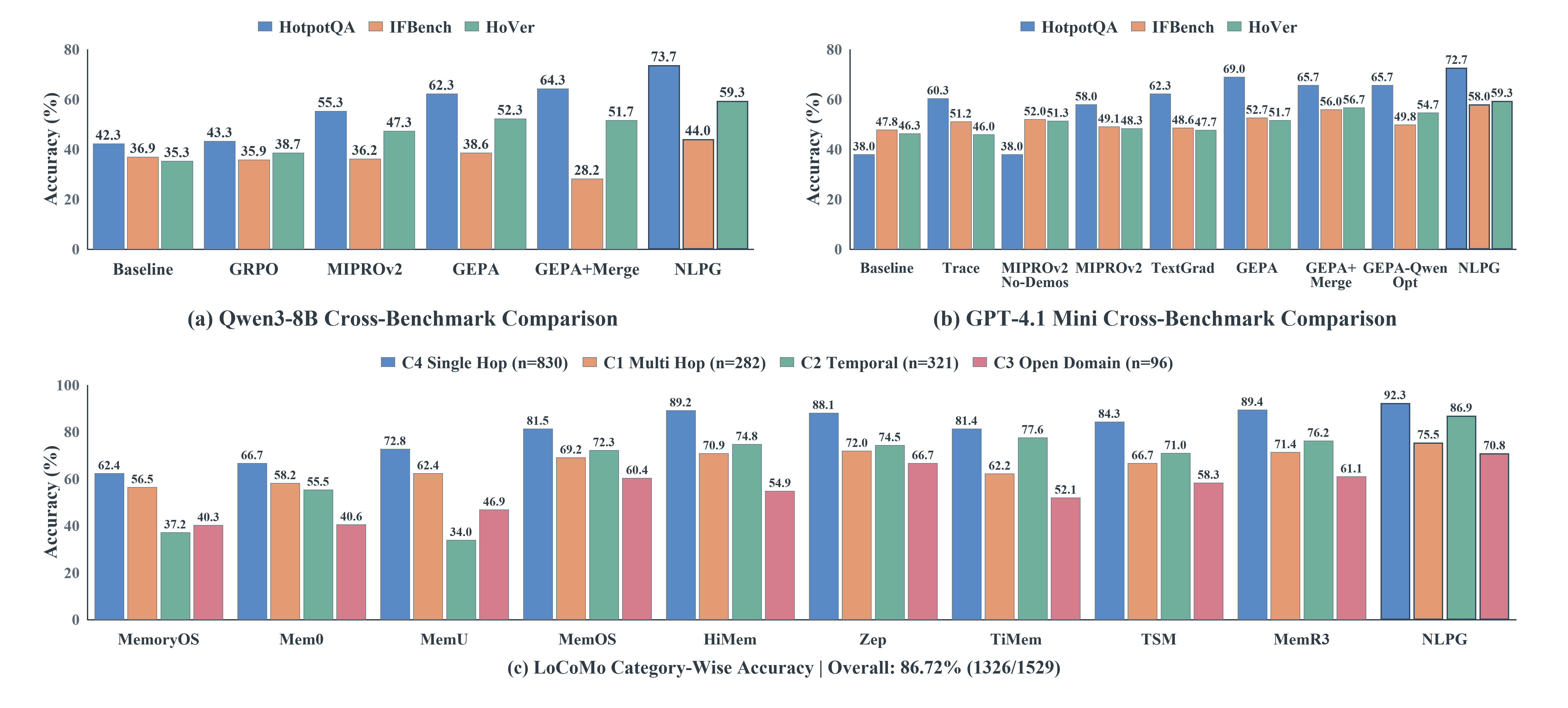}
    \caption{Cross-benchmark accuracy comparison. Panels (a) and (b) show transfer results for Qwen3-8B and GPT-4.1 Mini, respectively, while panel (c) reports category-wise LoCoMo accuracy. Consistent with Tables~\ref{tab:gepa_model_generalization} and~\ref{tab:locomo_results}, NLPG achieves the best performance on all transfer tasks for both backbones and improves every LoCoMo category, with the largest gains on multi-hop, temporal, and open-domain questions. The strongest improvements occur on procedurally compositional questions, suggesting that route-specific policy updates improve decisions about query preservation, memory combination, and retrieval expansion rather than simply increasing context size.}
    \label{fig:experiment-comparison}
\end{figure}

Figure~\ref{fig:experiment-comparison} summarizes the results. NLPG reaches
$86.72$ on LoCoMo, with the largest gains on multi-hop, temporal, and
open-domain questions; the latter category is small and should be interpreted
cautiously. It also obtains $96.20$ Memory Accuracy and $85.37$ QA Accuracy on
HaluMem, and $71.80$ QA Accuracy on LongMemEval. The transfer gains persist
across both backbones: on HotpotQA, IFBench, and HoVer, NLPG improves over the
corresponding baseline by $31.34,6.43,24.00$ points with Qwen3-8B and
$34.67,10.21,13.00$ points with GPT-4.1 Mini. Because NLPG improves tasks with different evidence-routing and instruction-following requirements using the same failure-to-correction update procedure, the result supports the portability of this procedure rather than the reuse of a single benchmark-specific answer format or policy text.

\subsection{Ablation Study}
\label{sec:ifbench-mechanism-ablation}

We ablate NLPG's core mechanisms and route-local memory capacity on the same 150-example IFBench split, holding the model, evaluation protocol, and inference budget fixed.

\begin{figure}[t]
    \centering
    \includegraphics[width=\linewidth]{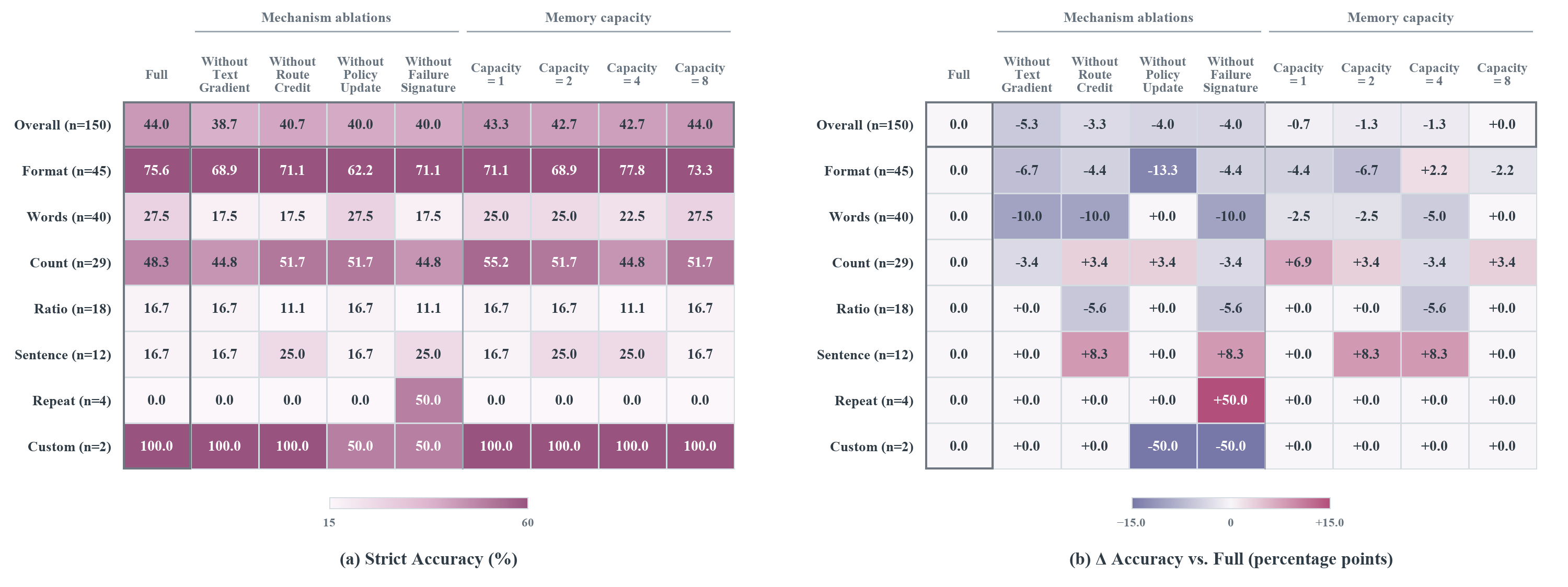}
    \caption{IFBench ablations by instruction category. \textbf{(a)} Strict
    accuracy (\%). \textbf{(b)} Difference from full NLPG in percentage
    points. Hatched cells contain fewer than 10 examples.}
    \label{fig:ifbench-ablation}
\end{figure}

Figure~\ref{fig:ifbench-ablation} summarizes the contribution of each
component. Removing textual gradients causes the largest drop $5.3\%$,
showing that natural-language corrections provide actionable guidance beyond
failure counts. Removing policy updates or failure signatures reduces accuracy by 4.0 percentage points, highlighting the importance of cross-example feedback
accumulation and explicit failure diagnosis. Removing route credit causes a further 3.3-point drop, providing evidence for the value of module-specific credit assignment.

\subsection{Analysis of Module-Local Policy Intervention}
\label{sec:policy_intervention}

\paragraph{Why specificity policies work.}
We compare the same case with and without NLPG's route-specific specificity
policy, keeping the question, backbone, retrieval setup, and QA program fixed.

\begin{tcolorbox}[
  enhanced,
  colback=gray!5,
  colframe=gray!60,
  colbacktitle=gray!60,
  coltitle=white,
  fonttitle=\bfseries,
  title=Case Study: NLPG Policy Intervention for Answer Specificity,
  arc=2mm,
  boxrule=0.6pt,
  left=4pt,
  right=4pt,
  top=4pt,
  bottom=4pt
]

\textbf{Question:} Trey Anastasio and Glenn Bidmead, have which mutual
occupation?

\textbf{Gold answer:} guitarist

\vspace{2mm}

\noindent
\begin{minipage}[t]{0.47\linewidth}
\raggedright
\textcolor{red!75!black}{\textbf{Without NLPG Policy}}\par
\vspace{1mm}
\textbf{Diagnosis:} answer overgeneralization\par
\vspace{5mm}
\textbf{Hop-2 query:}\par
What are the shared occupations of Trey Anastasio and Glenn Bidmead?\par
\vspace{1mm}
$\rightarrow$ \textbf{Predicted answer:} musician\par
\vspace{1mm}
\textcolor{red!75!black}{$\times$ Broader umbrella label}
\end{minipage}\hfill
\begin{minipage}[t]{0.47\linewidth}
\raggedright
\textcolor{green!50!black}{\textbf{With Specificity Policy}}\par
\vspace{1mm}
\textbf{Policy update:} preserve the most specific shared attribute\par
\vspace{1mm}
\textbf{Hop-2 query:}\par
What occupation do Trey Anastasio and Glenn Bidmead share?\par
\vspace{1mm}
$\rightarrow$ \textbf{Predicted answer:} guitarist\par
\vspace{1mm}
\textcolor{green!50!black}{$\checkmark$ Specific answer recovered}
\end{minipage}

\end{tcolorbox}

Supporting-document recall remains $1.0$ in both runs, so the repair does not
come from additional evidence. Instead, the specificity policy changes the
hop-2 query and replaces the broad label ``musician'' with the exact occupation
``guitarist'', consistent with the diagnosed overgeneralization error. As a
single case with multi-module injection, it neither isolates one module's
contribution nor provides a population-level causal estimate.
\section{Conclusion}
\label{sec:conclusion}

We introduced Natural-Language Policy Gradients (NLPG), a lightweight framework that converts execution failures into reusable, module-local natural-language policies for compound LLM agents. By combining failure diagnosis, route-aware policy updates, bounded selection, and offline validation, NLPG improves both conversational-memory and multi-hop reasoning tasks without changing model parameters or retrieval backends. Our results and ablations show that textual feedback and module-specific credit assignment are key to these gains, while a controlled intervention confirms that targeted policies can improve answer specificity under fixed retrieval. Future work will explore more automatic failure routing, larger-scale policy evolution, and extensions to multimodal and larger tool-using agents.

\clearpage

\subsection*{AI Use Statement}

AI-based assistants were used to support language polishing, retrieval and discovery, and code-review tasks during the preparation of this manuscript. All algorithmic designs, implementations, experiments, analyses, citations, and final manuscript content were checked and verified by the authors, who remain responsible for the accuracy and integrity of this work.

\subsection*{ETHICS STATEMENT}
NLPG improves LLM agents by extracting natural-language policies from prior
executions and applying them to subsequent tasks. Although this can improve
consistency and task performance, storing execution traces and derived
policies may introduce privacy risks when inputs contain sensitive information.
The learned policies may also preserve or amplify biases, errors, or
unsupported assumptions present in the data or model outputs. Therefore,
deployment should use appropriate safeguards, including user consent, data
minimization, anonymization, access control, policy validation, and human
oversight. We use public benchmark datasets solely for research evaluation and
do not recommend deploying NLPG for high-stakes decisions without additional
safety review.

\subsection*{REPRODUCIBILITY STATEMENT}

To support reproducibility, we describe the NLPG algorithm and execution
process in Section~3 and provide implementation details, prompt templates,
model assignments, data-separation protocols, benchmark adapters, and
evaluation scripts in Appendices~\ref{app:implementation} and
\ref{app:experimental-setup}. We report the model identifiers, random seeds,
data splits, policy-memory capacity, inference settings, scoring procedures,
and evaluation denominators used in the experiments. The policy-construction,
selection, and final-test partitions are disjoint, and the selected policy is
frozen during final evaluation. We plan to release the source code and
configuration files to facilitate independent verification and replication of
the reported results.

\bibliography{main}
\bibliographystyle{iclr2025_conference}

\clearpage
\appendix

\startcontents[appendix]

\section*{Appendix Contents}
\printcontents[appendix]{}{1}{%
  \setcounter{tocdepth}{3}
}

\section{Implementation Details}
\label{app:implementation}

NLPG adds an external procedural policy layer to a fixed compound agent. The
benchmark program remains responsible for memory construction, retrieval,
generation, and evaluation, while NLPG controls how reusable policy
instructions are selected, updated, and injected into the program. Model
parameters, program structure, retrieval indexes, and benchmark evaluators are
kept fixed. All reported NLPG results use the complete pipeline described
below; component removals are used only in the ablation experiments.

\subsection{Inference with a Frozen Policy}
\label{app:end-to-end-workflow}

During evaluation pass $t$, the agent processes each task instance using the
frozen policy memory $M_t$. The memory is indexed by execution routes, which may
correspond to task families, program modules, or a shared general route. For
the current route $a$, NLPG removes duplicate instructions, ranks the
remaining candidates, and injects at most $K$ instructions into the
corresponding program context. The fixed agent then executes its original
retrieval and generation procedure.

\begin{algorithm}[t]
\caption{Inference with a frozen route-indexed policy}
\label{alg:nlpg-inference}
\begin{algorithmic}[1]
\Require Task instance $x$, active policy memory $M_t$, budget $K$
\Ensure Response $\hat{y}$ and execution record $\tau$

\State $a \gets \operatorname{Route}(x)$
\State $\mathcal{P} \gets
\operatorname{TopK}\!\left(
\operatorname{Dedup}(M_t^{a}\cup M_t^{\mathrm{gen}}), K
\right)$
\State $\tau \gets \operatorname{Execute}(P_{\theta}, x, \mathcal{P})$
\State $\hat{y} \gets \operatorname{Response}(\tau)$
\State $z \gets \operatorname{Evaluate}(\tau)$
\State \Return $\hat{y},(\tau, z)$
\end{algorithmic}
\end{algorithm}

Here, $M_t^{\mathrm{gen}}$ is empty for benchmarks without a shared policy
route. The selected instructions are appended to the relevant module or task
context, while the ordering of the original program is unchanged. Each
execution record stores the observable inputs and outputs, route information,
retrieved evidence, policy instructions, execution dependencies, final
response, and evaluation result. The active policy is not modified while the
current evaluation pass is running.

\begin{table}[t]
\centering
\caption{Conceptual data flow of the NLPG execution and update process.}
\label{tab:nlpg-data-flow}
\small
\begin{tabularx}{\linewidth}{@{}l X X@{}}
\toprule
Stage & Input $\rightarrow$ Output & Role in NLPG \\
\midrule
Policy loading &
$M_t \rightarrow$ route-indexed policy candidates &
Load the active procedural policy version. \\

Policy selection &
Task input and route $\rightarrow$ top-$K$ instructions &
Deduplicate, rank, and select the policy context for the invocation. \\

Fixed execution &
Task input and selected policy $\rightarrow$ execution trace &
Run the original benchmark program without changing its structure. \\

Evaluation &
Execution trace $\rightarrow z_i$ &
Apply the benchmark's released checker or judge. \\

Failure analysis &
$(\tau_i, z_i) \rightarrow$ failure feedback &
Identify the deviation and the routes relevant to its correction. \\

Policy update &
Failure feedback $\rightarrow$ textual corrections &
Generate and consolidate reusable route-specific instructions. \\

Policy validation &
Candidate policy $\rightarrow$ accepted or rejected version &
Evaluate the candidate on a disjoint policy-selection set. \\
\bottomrule
\end{tabularx}
\end{table}

\subsection{Offline Policy Construction}
\label{app:offline-policy-update}

Policy construction takes place after an evaluation pass has finished. The
observed execution records are converted into failure feedback associated with
the routes that could have contributed to the deviation. When an execution
graph is available, feedback is propagated from terminal modules toward their
recorded predecessors in reverse topological order. The critic uses the local
execution context, module output, and downstream feedback to produce a
route-specific textual correction. Corrections from different executions are
then grouped by route and normalized instruction content.

\begin{algorithm}[t]
\caption{Route-aware offline policy construction}
\label{alg:nlpg-policy-update}
\begin{algorithmic}[1]
\Require Completed records $\mathcal{R}$ and active policy $M_t$
\Ensure Candidate policy $\widehat{M}_{t+1}$

\State $\mathcal{D} \gets \operatorname{Diagnose}(\mathcal{R})$
\State $\mathcal{G} \gets \operatorname{Critic}(\mathcal{D},M_t)$
\State $\mathcal{G}' \gets \operatorname{Consolidate}(\mathcal{G})$
\State $\mathcal{G}_{\mathrm{safe}} \gets
\operatorname{Filter}(\mathcal{G}')$
\State $\widehat{M}_{t+1} \gets
\operatorname{Merge}(M_t,\mathcal{G}_{\mathrm{safe}})$
\State \Return $\widehat{M}_{t+1}$
\end{algorithmic}
\end{algorithm}

For a grouped correction $u$ associated with route $a$ and failure signature
$s$, NLPG assigns the following retention score:
\begin{equation}
w_{a,s,u}
=
\frac{|\mathcal{E}_{a,s,u}|}
{\max(N_a,|\mathcal{E}_a|,1)}
\alpha_{a,s,u}\kappa_{a,s,u},
\label{eq:app-nlpg-candidate-score}
\end{equation}
where $\mathcal{E}_{a,s,u}$ is the set of distinct executions supporting the
correction, $N_a$ is the number of valid evaluated executions associated with
route $a$, and $\mathcal{E}_a$ is the set of executions contributing accepted
corrections on that route. The terms $\alpha_{a,s,u}$ and
$\kappa_{a,s,u}$ denote the normalized failure weight and critic confidence.
The score is used to rank candidates for retention; it is not interpreted as a
causal estimate of correction utility.

The safety stage removes empty, duplicate, overly specific, overlong,
label-leaking, or unsafe instructions and limits the number of retained
candidates per route. The resulting object is a candidate successor policy. It becomes active only after passing policy selection; otherwise, the current policy is retained for the subsequent evaluation pass.

\subsection{Benchmark Adapters and Policy Injection}
\label{app:adapter-contracts}

NLPG uses a common policy memory and update procedure across benchmarks. Since the evaluated programs expose different execution modules and evaluation signals, each benchmark is connected through an adapter that translates its execution trace and evaluation result into a common NLPG representation. This representation contains the evaluation signal, observed failure information, route metadata, and execution evidence required for policy improvement.

The adapter maps each benchmark's execution record and evaluator output to the
common NLPG representation, including the normalized evaluation signal, failure
descriptor, task family, and eligible policy routes. It therefore performs
benchmark-specific diagnosis and route assignment, but does not generate the
policy text or aggregate updates across executions. The shared NLPG critic
combines these normalized signals with the observable execution evidence to
produce reusable route-local corrections. The shared optimizer then groups and
ranks compatible corrections across executions, while the safety stage removes
duplicated, overly specific, label-leaking, or unsafe instructions before the
next policy version is constructed.

Route names follow the vocabulary of the fixed program. Memory-oriented
benchmarks generally expose task-family routes, whereas multi-hop and
instruction-following programs expose module- or constraint-specific routes.
This route conditioning allows feedback to be assigned at the level where the
corresponding behavior is executed, rather than pooling all corrections into a
single prompt.

\begin{table}[htbp]
\centering
\small
\caption{Execution routes and policy injection points used by the evaluated
benchmark programs.}
\label{tab:app-adapter-binding}
\begin{tabularx}{\linewidth}{@{}l X X c@{}}
\toprule
Benchmark & Route representation & Policy injection point & $K$ \\
\midrule
LoCoMo &
Question category &
Retrieval and answer guidance & 4 \\

HaluMem &
Inferred question family &
Memory retrieval and answer policy & 4 \\

LongMemEval &
Question type with fallback family &
Retrieval and answer guidance & 4 \\

HotpotQA &
Module routes in the two-hop program &
Summarization, second-hop query, and answer modules & 3 \\

IFBench &
Instruction identifier and instruction family &
Response-generation context & 3 \\

HoVer &
Hop-specific summary and query modules &
Multi-hop summary and query generation & 3 \\
\bottomrule
\end{tabularx}
\end{table}

Reference information is consumed by the benchmark evaluator and, where
required by the benchmark protocol, is converted into a normalized evaluation
signal or failure descriptor before policy construction. For example,
supporting-document annotations are used to compute document coverage in
HotpotQA and HoVer, while the official checker identifies unsatisfied
instruction constraints in IFBench. These evaluator-derived signals may
determine the failure signature and eligible policy routes. The critic does not
receive the raw gold answer, supporting-document annotations, or reference
text; it receives the observable execution record together with the normalized
evaluation feedback and failure descriptors. Policy instructions are generated
as reusable behavioral corrections.

\subsection{Policy Lifecycle and Held-Out Evaluation}
\label{app:policy-lifecycle}

NLPG maintains one active policy version during each evaluation pass. After the
pass, the collected records are used to construct a candidate successor
policy. In the held-out evaluation protocol, policy construction, policy
selection, and final testing use disjoint data partitions. The candidate and
its parent are compared under the same model configuration, program, and
evaluation criterion on the policy-selection partition. A candidate that
fails the required improvement or regression constraints is discarded; the
previous policy remains active.

After policy selection is completed, the adopted policy is evaluated once on
the final held-out test partition. Test executions do not contribute to policy
construction or policy selection. Each policy file records its version,
parent, route candidates, and the identifiers of the executions used to
construct and evaluate it, allowing partition overlap to be checked directly.

\subsection{Benchmark Execution and Retrieval}
\label{app:benchmark-implementation}

Model assignments are benchmark configurations rather than components of the
NLPG method. NLPG does not introduce additional model calls during ordinary
policy injection and does not alter the benchmark retrieval index. For HotpotQA, the fixed program performs two retrieval stages with intermediate summarization
and second-hop query generation before answer synthesis. NLPG is injected into
the language-generation modules only.

HoVer uses a fixed multi-hop retrieval program with hop-specific summaries and
queries. When a local corpus index is enabled, document retrieval is performed by the benchmark's configured lexical index, while language models
produce summaries and queries. Supporting-document coverage and document recall
are recorded as retrieval diagnostics, and the final verification score is
computed by the benchmark evaluator. Complete model assignments, retrieval
budgets, dataset partitions, and scoring rules are reported in
Appendix~\ref{app:experimental-setup}.

\subsection{Computational Overhead}
\label{app:notation-complexity}

For a route containing $C_a$ policy candidates, candidate ranking costs
$O(C_a\log C_a)$ and formatting the selected $K$ instructions costs
$O(KL)$, where $L$ is their mean length. This overhead is bounded by the
policy budget and adds no mandatory retrieval operation.

Offline construction is linear in the number of inspected execution records
and generated corrections, followed by route-wise candidate ranking. Critic
and graph-feedback operations add language-model calls only during policy
construction. The main configuration bounds the number and length of retained
instructions and corrections, ensuring that the policy context remains
bounded across evaluation passes.

\FloatBarrier
\section{Related Work}
\label{sec:related_work}

\paragraph{Compound agents.}
Compound language agents coordinate multiple language-model calls with
retrieval, tool use, and verification. ReAct interleaves reasoning and acting
\citep{yao2023react}, Toolformer studies tool use by language models
\citep{schick2023toolformer}, AutoGen supports multi-agent coordination
\citep{wu2023autogen}, and DSPy represents language-model applications as
composable programs \citep{khattab2024dspy}. These works establish the
execution structure of compound agents, but, to our knowledge, have paid less attention to explicit post-deployment, module-level policy updates in compound agents.

\paragraph{External memory.}
Memory-augmented agents preserve information across interactions through
experience records, reflection, or hierarchical storage
\citep{park2023generative,packer2023memgpt}. Mem0, LightMem, and ProMem
further improve memory extraction, compression, retrieval, and verification
\citep{chhikara2025mem0,fang2025lightmem,yang2026promem}. These methods mainly
manage historical content or factual memories. NLPG instead stores reusable
procedural instructions indexed by task family or program module and updates
them from diagnosed execution failures.

\paragraph{Language-based agent optimization.}
Reflexion and Self-Refine use verbal feedback to improve later attempts
\citep{shinn2023reflexion,madaan2023selfrefine}. Prompt-optimization methods
search over instructions through textual edits, language-model optimization,
or evolutionary search \citep{pryzant2023apo,yang2024opro,
opsahlong2024miprov2,fernando2024promptbreeder}. TextGrad formalizes language
feedback as a textual gradient, while Trace uses execution traces and rich
feedback for generative optimization \citep{yuksekgonul2025textgrad}. GEPA extends this direction to reflective prompt evolution
for compound systems \citep{agrawal2026gepa}.

\section{Experimental Details}
\label{app:experimental-setup}

This appendix records the experimental settings, model assignments, data flow,
and evaluation protocols for NLPG. Comparisons within one block use the same
dataset version, final-test examples, task interface, candidate corpus,
retrieval budget, answer format, evaluator, normalization rule, and scoring
denominator.

All NLPG results are evaluated on an independently held-out final set. Before
policy construction, each benchmark is partitioned into a construction set, a
selection set, and a final test set. Evaluated trajectories from the
construction set are used to produce candidate policies, whereas the selection
set is used only to choose between candidate and parent policy versions. The
selected policy is then frozen and evaluated once on the final test set. Final-
test inputs, outputs, labels, and evaluator feedback are never used to construct,
select, revise, or rescore the policy reported for that set.

\subsection{Data Separation and Result Provenance}
\label{app:data-separation}

Table~\ref{tab:app-data-flow} summarizes the separation protocol. Splits are
materialized before optimization and stored with stable example identifiers.
For benchmarks containing multiple questions about the same conversation,
user, memory record, or document history, separation is performed at that
higher-level unit rather than only at the question level. Consequently, no
underlying context unit can contribute feedback to policy construction or
selection and later reappear in the final test set.

\begin{table}[htbp]
\centering
\small
\caption{Independent held-out evaluation protocol. Construction, selection,
and final-test partitions are mutually disjoint under the indicated grouping
key.}
\label{tab:app-data-flow}
\begin{tabularx}{\linewidth}{@{}lXXXX@{}}
\toprule
Benchmark & Policy construction & Policy selection & Final test & Separation key \\
\midrule
LoCoMo & Non-test conversation--user groups & Disjoint conversation--user groups & 152 first-user questions from held-out groups & Conversation and user ID \\
HaluMem & Non-test user records & Disjoint validation user records & 164 questions from held-out user records & User and record ID \\
LongMemEval & Non-test history--question groups & Disjoint validation groups & Held-out LongMemEval-S questions & History and question ID \\
HotpotQA & 150 train examples & 300 validation examples & 300 test examples & Example ID \\
IFBench & Construction portion of calibration data & Disjoint selection portion & 150 held-out examples & Prompt/example ID \\
HoVer & Construction portion of 150 train examples & Disjoint selection portion & 300 test examples & Claim ID \\
\bottomrule
\end{tabularx}
\end{table}

The same final-test manifest is used for every method in a comparison block.
Published baseline values are reused only when their split and evaluation
protocol match that manifest; otherwise, the baseline is rerun on the shared
final set. Split manifests are fixed before any policy update, and overlap is
checked using both example identifiers and the benchmark-specific grouping keys
listed in Table~\ref{tab:app-data-flow}.

\subsection{Benchmark and Model Alignment}
\label{app:benchmark-alignment}

We evaluate NLPG on six benchmarks covering conversational memory, faithful
memory construction, long-term memory question answering, and general
compound-agent optimization. The memory benchmarks are LoCoMo
\cite{maharana2024locomo}, HaluMem \cite{chen2025halumem}, and LongMemEval
\cite{wu2025longmemeval}. The transfer benchmarks are GEPA-aligned HotpotQA
\cite{yang2018hotpotqa}, IFBench \cite{pyatkin2025ifbench}, and HoVer
\cite{jiang2020hover}.

All results are organized into comparison blocks. A comparison block is
defined by a benchmark, a fixed dataset subset, a task interface, and a
specified model assignment. Results from different blocks are not merged when
their backbone, evaluator, or runtime protocol differs.

For LoCoMo, the shared comparison uses the same held-out first-user questions
and the same memory-building, answer-generation, and judging configuration for
every method in the corresponding table. No conversation or user record from
this final partition is used during policy construction or selection.

For HotpotQA, IFBench, and HoVer, each model block uses one fixed task
backbone for all task modules in that block. Auxiliary models used for
reflection, criticism, or policy optimization are reported separately and are
not treated as the task backbone. In particular, a model used by an auxiliary
critic or optimizer is not used to relabel the backbone of the corresponding
task result.

\subsection{LoCoMo}
\label{app:locomo-details}

LoCoMo evaluates question answering over long conversational histories~\cite{maharana2024locomo}. We use the complete released LoCoMo evaluation set, including all valid questions and the corresponding conversation histories, user identifiers, and question categories. Every method receives the same conversation history and question and produces an answer through the same answer interface.

The NLPG policy is constructed and selected using conversation--user groups disjoint from the final evaluation partition. The selected policy is frozen before evaluation on the complete held-out LoCoMo test set. Conversation, user, and question identifiers are checked for overlap before scoring, and final-test trajectories and judgments do not enter any subsequent policy update.

The NLPG policy is constructed and selected using conversation--user groups
that are disjoint from the final evaluation partition. The selected policy is
then frozen and evaluated once on the 152 held-out first-user questions from
categories 1--4. Conversation, user, and question identifiers are checked for
overlap before scoring; trajectories and judgments from these 152 questions do
not enter a subsequent update of the reported policy.

The comparison includes MemoryOS, Mem0 \cite{chhikara2025mem0}, MemU, MemOS,
HiMem, Zep, TiMem, TSM, MemR3, and NLPG. Baseline values are reused only when
the published or released configuration matches the comparison protocol.
Otherwise, the method is evaluated in the shared harness using the same
questions, answer format, evaluator, and denominator.

The LoCoMo table uses the model assignment specified by the comparison block.
Memory construction, answer generation, and judging use the same corresponding
models for all methods, and only the method-specific memory or policy mechanism
is changed.

LoCoMo accuracy is computed as the proportion of valid questions judged
correct. Percentages are reported at the precision used by the corresponding
result table. When a source reports a rounded percentage, the value is
retained as reported rather than being reverse-converted into an integer count.

\subsubsection{Paired Ablation Analysis on LoCoMo}
\label{sec:locomo-paired-ablation}

\begin{figure}[t]
    \centering
    \includegraphics[width=\linewidth]
    {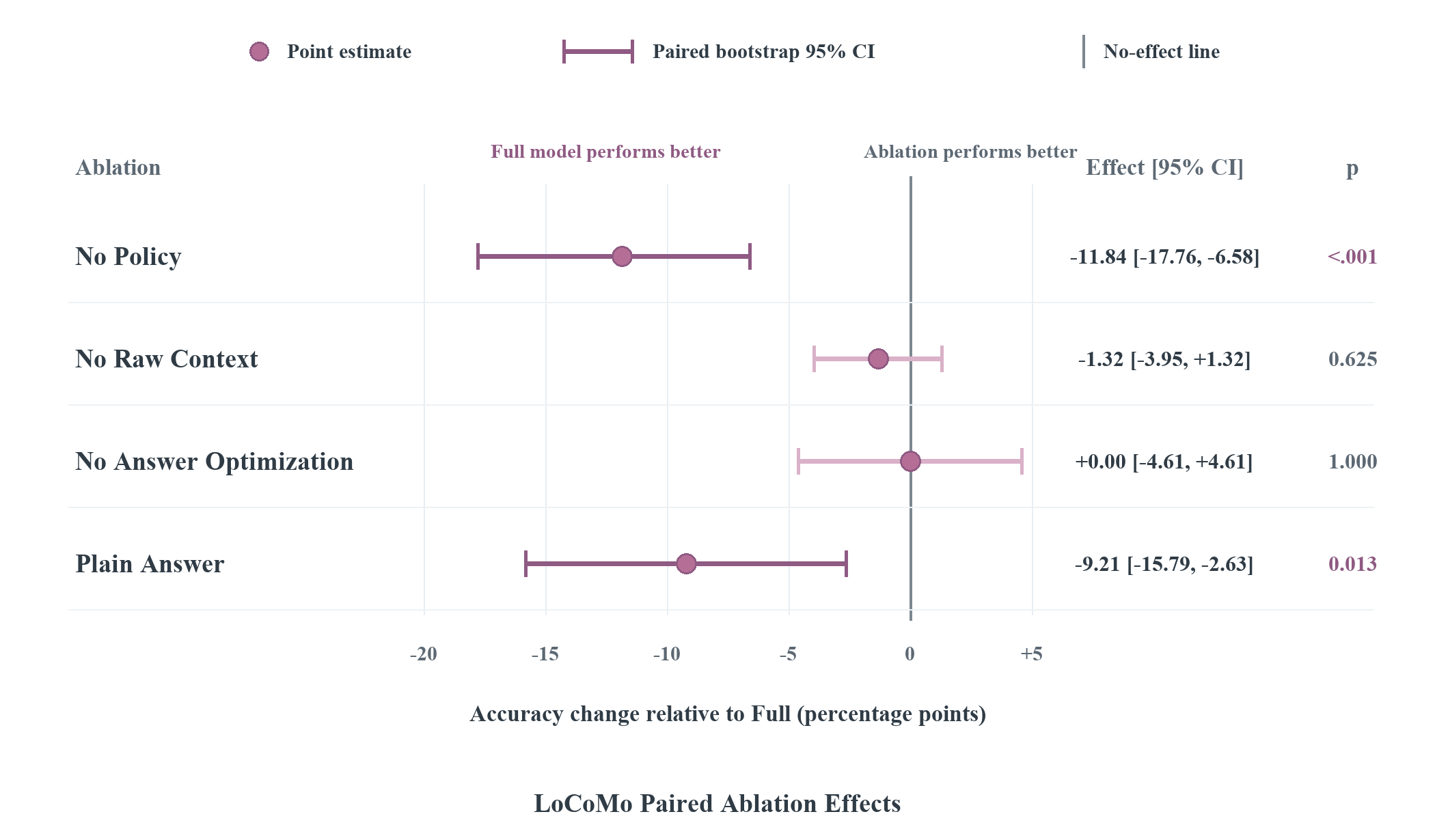}
    \caption{Paired answer-stage ablations on LoCoMo
    ($n=152$). Points denote accuracy changes relative to the full
    model, and horizontal bars show paired-bootstrap 95\% confidence
    intervals. Negative values favor the full model, while the vertical
    line denotes no effect. Reported $p$-values are from two-sided exact
    McNemar tests. Retrieval results are fixed and cached across all
    variants.}
    \label{fig:locomo-paired-ablation}
\end{figure}

We further conduct paired ablations on the same 152 held-out LoCoMo questions
under fixed cached retrieval. As shown in
Figure~\ref{fig:locomo-paired-ablation}, removing  the policy causes a
significant accuracy drop of 11.84 percentage points, while replacing
the optimized answer stage with plain answering reduces accuracy by
9.21 points. In contrast, removing raw context or answer optimization
alone does not yield a statistically significant difference. These
results highlight the importance of policy and
the complete answer-generation design.

\subsubsection{Category-level Analysis of Answer-stage Ablations on LoCoMo}
\label{sec:locomo-category-ablation}

\begin{figure}[t]
    \centering
    \includegraphics[
        width=\linewidth
    ]{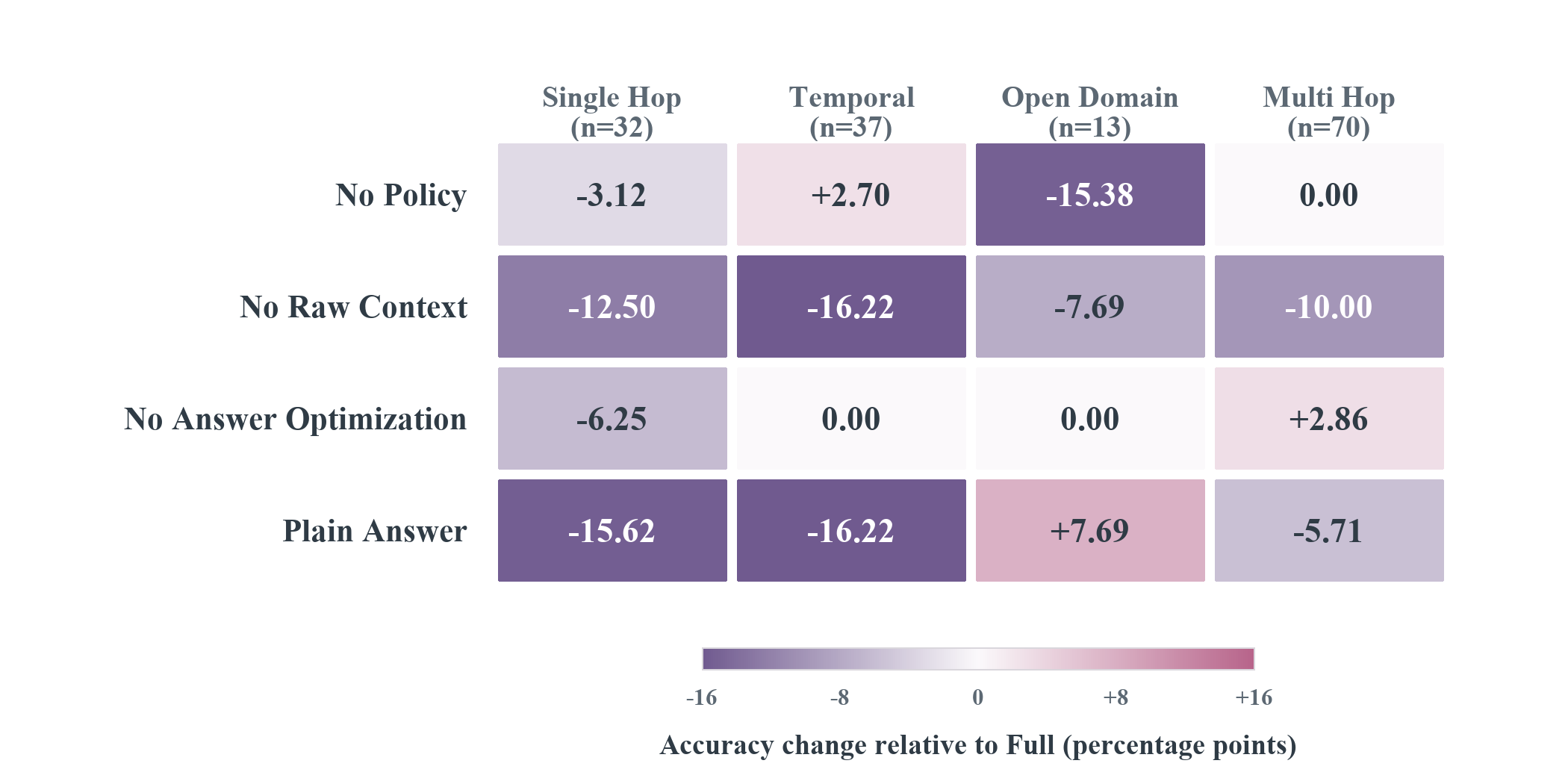}
    \caption{
    Category-level changes in accuracy, measured in percentage points relative to the full configuration, for answer-stage ablations on the held-out LoCoMo set with fixed cached retrieval. Negative values indicate that the ablated configuration performs worse than the full configuration. The numbers in parentheses denote the number of questions in each category. The Open Domain column is interpreted cautiously because this category contains only 13 questions.
    }
    \label{fig:locomo-category-ablation}
\end{figure}

To examine whether answer-stage components affect different question types differently, we report category-level accuracy changes relative to the full configuration under fixed cached retrieval. As shown in Figure~\ref{fig:locomo-category-ablation}, removing the raw conversational context causes the most consistent degradation across all categories, with drops ranging from 7.69 to 16.22 percentage points. Removing answer optimization produces comparatively small changes, while the effects of removing policy prompts are mixed across categories. The Open Domain results should be interpreted cautiously because this category contains only 13 questions.

\subsection{Faithful Memory Construction}
\label{app:halumem-details}

HaluMem jointly evaluates the integrity and factual accuracy of constructed
memories and their downstream usefulness for question answering
\cite{chen2025halumem}. We use the released examples, memory targets, and
downstream questions. All methods in the comparison use the same examples,
memory targets, questions, answer interface, evaluator, and scoring
denominator.

HaluMem is partitioned by user record before policy learning. Policy
construction and selection use disjoint non-test users, and the selected policy
is evaluated on 164 questions associated only with held-out users. This
user-level partition prevents the same profile, conversation history, memory
target, or downstream question from appearing in more than one data role.

The three reported metrics are Memory Integrity, Memory Accuracy, and
downstream QA Accuracy. Memory Integrity measures whether the constructed
memory preserves the required information without introducing invalid content.
Memory Accuracy evaluates the correctness of the stored memory relative to the
benchmark target. QA Accuracy measures whether the system answers downstream
questions correctly using the constructed memory.

\begin{table}[htbp]
\centering
\small
\caption{Results on HaluMem under the aligned benchmark evaluation protocol.
The three metrics are reported separately because they measure different
properties of memory construction and downstream use.}
\label{tab:halumem_results}
\setlength{\tabcolsep}{8pt}
\begin{tabular}{lccc}
\toprule
Method & Memory Integrity & Memory Accuracy & QA Accuracy \\
\midrule
MemoBase & 14.55 & 92.24 & 35.53 \\
Supermemory & 41.53 & 90.32 & 54.07 \\
Mem0 & 42.91 & 86.26 & 53.02 \\
ProMem & \textbf{73.80} & 89.47 & 62.26 \\
NLPG & 70.25 & \textbf{96.20} & \textbf{85.37} \\
\bottomrule
\end{tabular}
\end{table}

On the independent held-out set, NLPG obtains the highest Memory Accuracy and
QA Accuracy among the listed values, while ProMem obtains the highest Memory
Integrity.

\subsection{Long-Term Memory Question Answering}
\label{app:longmemeval-details}

LongMemEval evaluates question answering over extended interaction histories
in which relevant evidence may be distributed across multiple sessions
\cite{wu2025longmemeval}. We use the same histories, questions, valid-question
denominator, answer interface, evaluator, and accuracy definition for all
methods in the aligned comparison. The comparison follows the protocol used
for NativeRAG, Mem0, LightMem, and ProMem.

LongMemEval is partitioned before policy learning using both question and
history identifiers. Construction and selection operate on mutually disjoint
non-test groups, and the frozen selected policy is evaluated only on the
held-out LongMemEval-S groups. All baselines are scored on the same final-test
manifest and denominator.

\begin{table}[htbp]
\centering
\small
\caption{Results on LongMemEval under the aligned question-answering
protocol.}
\label{tab:longmemeval_results}
\setlength{\tabcolsep}{10pt}
\begin{tabular}{lc}
\toprule
Method & QA Accuracy \\
\midrule
NativeRAG & 65.09 \\
Mem0 & 43.31 \\
LightMem & 68.64 \\
ProMem & 69.57 \\
NLPG & \textbf{71.80} \\
\bottomrule
\end{tabular}
\end{table}

NLPG is 2.23 points above ProMem and 3.16 points above LightMem on the shared
held-out evaluation set.

\subsection{GEPA-Aligned HotpotQA}
\label{app:gepa-details}

\paragraph{Dataset construction.}

To match the referenced GEPA task setting, we load the train split of the
HuggingFace HotpotQA fullwiki collection as the base pool. Following the
shared splitter, the first 40\% is used as test, the next 40\% as validation,
and the final 20\% as train. A random seed of 1 is then used to retain 300
test examples, 300 validation examples, and 150 train examples.

This construction aligns the data partition and example counts with the
referenced GEPA setting. Each method is evaluated with the same
per-example corpus and two-hop program interface within the corresponding
comparison block.

The 150-example train split is used exclusively for policy construction, the
300-example validation split is used for policy selection, and the selected
policy is frozen before a single pass over the 300-example test split. No test
trajectory, gold answer, supporting-document annotation, or evaluator outcome
is available to the critic or optimizer before the final score is recorded.

\paragraph{Runtime configuration.}

The reported main run evaluates 300 test examples. Retrieval uses top-$k=5$
for both hop 1 and hop 2. The summary, retrieval, and answer modules use the
backbone specified by the relevant model block. Time filtering and explicit
thinking are disabled. NLPG selects at most $K=3$ policy instructions per
module, subject to a maximum policy-context length of 1,000 characters.

\begin{table}[htbp]
\centering
\small
\caption{Main configuration of the 300-example HotpotQA evaluation.}
\label{tab:app-hotpotqa-config}
\begin{tabularx}{\linewidth}{@{}lX@{}}
\toprule
Setting & Value \\
\midrule
Dataset & HotpotQA fullwiki, GEPA-aligned split policy \\
Evaluation split & Test; 300 examples; subset seed 1 \\
Data partition & 40\% test, 40\% validation, 20\% train \\
Retrieval & Two-hop retrieval; top-$k=5$ for each hop \\
NLPG policy context & At most 3 instructions per module; maximum 1,000 characters \\
Time filtering & Disabled \\
Explicit thinking & Disabled \\
Primary metric & Exact match after lowercasing, punctuation normalization, and article removal \\
Auxiliary diagnostics & Answer containment, full supporting-document coverage, and mean document recall \\
\bottomrule
\end{tabularx}
\end{table}

\paragraph{Metrics and diagnostics.}

For an answer to be counted as correct, the normalized prediction must exactly
match the normalized gold answer. Document recall is computed over the unique
gold supporting titles. Full supporting-document coverage requires every gold
supporting title to appear in the deduplicated two-hop retrieval result.

On the 300-example held-out test set, the frozen selected policy obtains 221
exact matches, corresponding to 73.67\% accuracy. The same run obtains 89.67\%
full-document coverage and 94.33\% mean document recall.

The offline two-hop analysis further separates retrieval failures from answer
synthesis failures. Among 82 examples requiring a second-hop target, 31
examples miss the second-hop page, while 51 recover all missing targets.
Separately, 59 examples are incorrect despite complete supporting-document
coverage, indicating an answer-synthesis error rather than a retrieval miss.

\subsection{IFBench and HoVer}
\label{app:ifbench-hover-details}

IFBench evaluates instruction-following behavior using its official task
construction and strict checker \cite{pyatkin2025ifbench}. We use the released
examples and evaluate all methods with the same instruction format, output
interface, checker, and scoring denominator. A prediction is counted as correct
only when it satisfies the constraints required by the official checker.

For NLPG, the calibration data are partitioned into disjoint construction and
selection subsets before optimization. A separate 150-example held-out split is
used only for final evaluation. The three partitions contain no shared prompt
or example identifiers, and feedback from the held-out pass is not used to
revise or rescore the reported policy.

HoVer evaluates evidence-grounded multi-hop verification
\cite{jiang2020hover}. We use the released claims, evidence requirements, and
verification protocol. The evidence retrieval, summary, and verification
interfaces are fixed within each model block. If an auxiliary model is used
for policy diagnosis or optimization, it is recorded separately from the
task-model assignment.

The 150-example training subset is divided into mutually disjoint construction
and selection portions. The selected NLPG policy is then frozen for a
300-example test subset. Claim identifiers have zero overlap across the three
partitions, and no failure labels or trajectories from the test subset are used
to select, rewrite, or rescore the reported policy.

For both benchmarks, the comparison includes the task baseline, GRPO,
MIPROv2 \cite{opsahlong2024miprov2}, TextGrad \cite{yuksekgonul2025textgrad},
Trace/OptoPrime \cite{cheng2024trace}, GEPA
\cite{agrawal2026gepa}, and the corresponding reported variants where
applicable. All methods within a model block use the same task backbone and
task interface.

\subsubsection{Feedback Quality and Repair Analysis on IFBench}
\label{app:ifbench-feedback-quality}

We evaluate whether NLPG feedback correctly identifies failed constraints and
whether the resulting instructions can repair an unsuccessful response. We use
the calibration split of IFBench, which contains 150 examples, and evaluate all
responses using the official strict and loose IFBench checkers. The response
model is Qwen3-8B, with temperature set to $0$ and a maximum generation length
of 512 tokens. Diagnoses are produced by the complete Agentic-NLPG pipeline,
including graph-based backward attribution, structured failure diagnosis,
evidence grounding, route-level credit assignment, and policy optimization.

\paragraph{Diagnosis Quality}
\label{app:ifbench-diagnosis-quality}

The diagnosis audit contains 98 matched records. We evaluate a diagnosis against
the constraints that the official strict checker identifies as failed. The
exact-signature Macro F1 is 73.81\%, while the hierarchical-signature Macro F1
is 75.85\%. Hierarchical matching treats a specific diagnosis as a valid
refinement of its official parent constraint. For example,
\textbf{count:punctuation:missing\_interrobang} is treated as a refinement of
\textbf{count:punctuation}. The hierarchical-signature Micro F1 is 74.14\%,
and the route-family Macro F1 is 82.14\%.

Beyond signature matching, 77.55\% of the audited diagnoses are grounded in an
observed failed constraint. The optimizer retains 43 distinct diagnostic
signatures, of which 88.37\% are grounded in the corresponding execution
evidence. These results indicate that NLPG generally attributes a failure to
the correct constraint family, even when the generated signature is more
specific than the identifier used by the official checker.

\paragraph{Official Paired Repair Evaluation}
\label{app:ifbench-paired-repair}

To measure whether the generated feedback is actionable, we provide the
NLPG-generated diagnosis instructions, the original request, and the original
response to the same Qwen3-8B response model. The model is asked to produce a
revised response, which is then evaluated using the official IFBench checkers.
No approximate or model-based verifier is used for the reported repair results.

We distinguish between a failure-only analysis and a mixed paired analysis.
The failure-only subset contains 91 examples for which NLPG produces a usable
diagnosis. All 91 examples fail the strict checker before repair, while five
already pass the loose checker. Therefore, the failure-only strict regression
rate is undefined rather than zero.

For the regression-aware paired evaluation, we combine the 91
diagnosis-driven failures with 52 responses that pass the strict checker before
repair. For the latter responses, the repair prompt contains only a
preservation instruction that asks the model to retain all already satisfied
constraints. Seven additional failed examples have no usable diagnosis and
are therefore excluded from the repair analysis. The resulting mixed paired
set contains 143 examples.

\begin{table}[t]
\centering
\small
\setlength{\tabcolsep}{5pt}
\renewcommand{\arraystretch}{1.08}
\caption{Official paired repair evaluation on the mixed IFBench calibration
subset. Repair success is computed over responses that fail the corresponding
checker before repair. Regression is computed over responses that pass the
corresponding checker before repair.}
\label{tab:ifbench-feedback-repair}
\begin{tabular}{@{}lrr@{}}
\toprule
Metric & Strict & Loose \\
\midrule
Paired records
    & 143
    & 143 \\
Pass before repair
    & 52/143 (36.36\%)
    & 57/143 (39.86\%) \\
Pass after repair
    & 59/143 (41.26\%)
    & 64/143 (44.76\%) \\
Repair success
    & 8/91 (8.79\%)
    & 8/86 (9.30\%) \\
Regression
    & 1/52 (1.92\%)
    & 1/57 (1.75\%) \\
Net gain
    & $+7$
    & $+7$ \\
Failed constraints before $\rightarrow$ after
    & $97 \rightarrow 89$
    & $91 \rightarrow 84$ \\
\bottomrule
\end{tabular}
\end{table}

As shown in Table~\ref{tab:ifbench-feedback-repair}, the strict pass rate
increases from 36.36\% to 41.26\%, while the loose pass rate increases from
39.86\% to 44.76\%. Under the strict checker, NLPG repairs 8 of the 91
initially failed responses and causes one of the 52 initially successful
responses to fail. This produces a net gain of seven successful responses.
The loose checker yields the same net gain, with eight successful repairs and
one regression. The number of failed individual constraints also decreases
from 97 to 89 under strict evaluation and from 91 to 84 under loose evaluation.

These results show that NLPG feedback yields a positive paired improvement,
although the end-to-end repair rate remains modest. The low regression rate
indicates that the preservation instruction usually retains already satisfied
constraints, but the single observed regression also shows that repair is not
risk-free. We therefore report repair gain and regression separately rather
than presenting only the post-repair accuracy.

\begin{figure}[t]
\centering
\includegraphics[width=0.98\linewidth]
{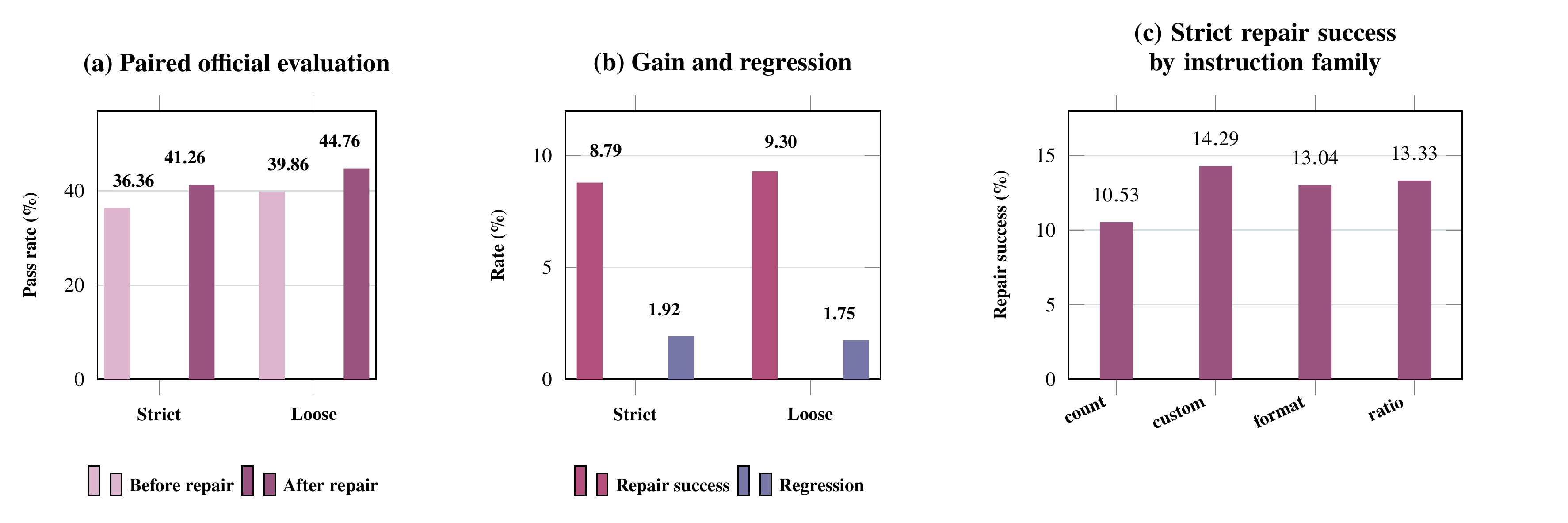}
\caption{Feedback quality and repair evaluation on IFBench. (a) Paired official evaluation, showing pass rates before and after repair under strict and loose checkers. (b) Gain and regression, showing repair-success and regression rates. (c) Strict repair success by official instruction family.}
\label{fig:ifbench-feedback-summary}
\end{figure}

\paragraph{Constraint-Family Analysis}
\label{app:ifbench-constraint-family}

Figure~\ref{fig:ifbench-feedback-summary} groups the diagnosis-driven failures
by the prefix of the official IFBench instruction identifier. An example can
contribute to multiple families when multiple constraints fail simultaneously.
For each family, the denominator is restricted to examples containing that
family that fail the strict checker before repair.

The highest observed repair success is obtained for the
\textbf{custom} family, with 1 of 7 examples repaired (14.29\%). The
\textbf{ratio} family achieves 2 of 15 repairs (13.33\%), the
\textbf{format} family achieves 3 of 23 repairs (13.04\%), and the
\textbf{count} family achieves 2 of 19 repairs (10.53\%). No complete
failure-to-pass repair is observed for the evaluated \textbf{repeat},
\textbf{sentence}, or \textbf{words} examples.

The family-level results highlight a distinction between diagnosis quality and
response-level repair success. NLPG can identify and ground a failed constraint
even when the response model cannot execute the corresponding correction
reliably. This distinction is particularly important for word-level and
sentence-level constraints, which often require a complete response to satisfy
a global lexical or structural property. Consequently, diagnosis F1 and
end-to-end repair success measure complementary aspects of feedback quality.

\paragraph{Repair-Prompt Control}
\label{app:ifbench-repair-prompt-control}

We additionally compare two repair-prompt variants. The \textbf{basic} variant
contains the original request, the unsuccessful response, and the
NLPG-generated diagnosis instructions. The \textbf{checklist} variant also
exposes the official failed instruction identifiers and explicitly asks the
model to verify every failed constraint while preserving constraints that the
original response already satisfies.

On the failure-only subset, the basic prompt repairs 8 of 91 responses under
strict evaluation, corresponding to 8.79\%. The checklist prompt repairs 7 of
91 responses, corresponding to 7.69\%. Under loose evaluation, the basic and
checklist variants produce 13 and 12 passing responses after repair,
respectively.

The checklist prompt therefore does not improve over the basic diagnosis
prompt in this run. This result suggests that the main bottleneck is not merely
the absence of an explicit instruction identifier. Instead, the remaining
difficulty lies in executing multiple simultaneous output constraints with the
response model. For this reason, we use diagnosis matching, evidence grounding,
repair success, and regression as separate measurements of NLPG feedback
quality.

\subsection{Evaluation Metrics}
\label{app:evaluation-metrics}

LoCoMo and LongMemEval are evaluated using benchmark question-answering
accuracy. HaluMem reports Memory Integrity, Memory Accuracy, and downstream QA
Accuracy under the definitions provided by the benchmark. HotpotQA uses
lowercased, punctuation-normalized exact match with articles removed. IFBench
uses the official strict instruction-following checker. HoVer uses its
evidence-grounded verification score.

All percentages are reported in percentage points. When an auxiliary metric is
computed, its denominator is fixed to the valid examples of the corresponding
benchmark split. For the three-task transfer average, we compute the arithmetic
mean of the HotpotQA, IFBench, and HoVer scores within the same model block.

\subsection{Policy-Memory Runtime}
\label{app:policy-runtime}

NLPG keeps the compound-agent program and underlying model parameters fixed
while updating an external policy memory. A policy entry contains a short
natural-language instruction, a utility score, and the identifiers of the
examples that support the instruction. During execution, the relevant
module-local policy entries are ranked by utility, deduplicated, and truncated
to the configured top-$K$ budget before being inserted into the module context.

The policy version used for an evaluation pass is frozen throughout that pass.
After evaluation, failure traces are grouped by failure signature and routed
to the relevant task category or program module. The resulting update
directions are aggregated across examples, and only updates satisfying the
configured support and utility criteria are retained for the next policy
version.

This separation ensures that reported task performance is measured with a
fixed active policy and that policy construction does not alter the model
weights, retrieval corpus, or program execution order during the same
evaluation pass.

\subsection{Controlled Policy Intervention}
\label{app:controlled-intervention}

We additionally conduct a paired intervention on one successful repair case.
The question, backbone models, corpus, two-hop program, retrieval budget, and
evaluation procedure are held fixed. Only a module-local specificity policy is
added in the second run.

The failed trace has complete supporting-document recall but predicts the
broader occupation ``musician'' instead of the gold answer ``guitarist''. We
therefore diagnose an answer-overgeneralization error and construct a targeted
instruction requiring the intersection of the two persons' explicitly stated
occupations. The instruction is routed to the summary, second-hop query, and
answer modules before rerunning the same example.

\begin{table}[htbp]
\centering
\small
\caption{Successful controlled policy intervention. Retrieval coverage is
unchanged; only the module-local policy differs between the paired runs.}
\label{tab:app-controlled-intervention}
\begin{tabularx}{\linewidth}{@{}lXX@{}}
\toprule
Observable & Without policy & With policy \\
\midrule
Question &
\multicolumn{2}{l}{Trey Anastasio and Glenn Bidmead have which mutual occupation?} \\
Supporting-document recall & 1.0 & 1.0 \\
Hop-2 query &
What are the shared occupations of Trey Anastasio and Glenn Bidmead? &
What occupation do Trey Anastasio and Glenn Bidmead share? \\
Predicted answer & musician & guitarist \\
Exact-match correctness & false & true \\
\bottomrule
\end{tabularx}
\end{table}

The injected answer-module instruction is:

\begin{quote}
Intersect the occupation attributes stated for both people and return the most
specific shared answer span, rather than a broader umbrella term.
\end{quote}

This intervention is a selected case study rather than an estimate of
aggregate accuracy improvement. It provides controlled evidence that a
module-local policy can repair answer specificity while leaving retrieval
coverage unchanged.

\FloatBarrier

\section{Prompt Templates}
\label{app:prompts}

This appendix describes the prompts used by NLPG. The fixed application
prompts define the input and output interfaces of the program modules. The
policy-update prompts operate on evaluated execution traces and generate
module-local procedural corrections. The model parameters, program structure,
retrieval interface, and evaluation procedure remain fixed during policy
evolution.

\newtcolorbox{nlpgprompt}[1]{
  title={#1},
  colback=black!6,
  colframe=black,
  colbacktitle=black,
  coltitle=white,
  fonttitle=\bfseries,
  boxrule=0.8pt,
  arc=2mm,
  outer arc=2mm,
  left=3mm,
  right=3mm,
  top=2mm,
  bottom=2mm,
  before skip=7pt,
  after skip=10pt
}

\subsection{Application-Level Prompts}
\label{app:application-prompts}

The application prompts specify the information consumed and produced by each
module. They provide the execution interface without prescribing
benchmark-specific failure corrections.

\begin{nlpgprompt}{Evidence Summarization}
You are the evidence summarization module.

Summarize the retrieved passages with respect to the current question and
retain information that may be required by subsequent modules.

Return a structured response containing a concise summary and the salient
entities, relations, or constraints found in the evidence.

\textbf{Question:} \{question\}

\textbf{Retrieved passages:}

\textbf{[1] \{title\_1\}}\\
\{text\_1\}

\textbf{[2] \{title\_2\}}\\
\{text\_2\}

\hspace*{1em}$\cdots$
\end{nlpgprompt}

\begin{nlpgprompt}{Query Generation}
You are the query-generation module.

Generate the next retrieval query from the task input and the preceding
module output.

Return the query together with a concise rationale.

\textbf{Question:} \{question\}\\
\textbf{Previous module output:} \{previous\_output\}
\end{nlpgprompt}

\begin{nlpgprompt}{Answer Synthesis}
You are the answer-synthesis module.

Use the retrieved evidence and intermediate outputs to produce the final answer.
Return the answer and the evidence identifiers used by the program.

\textbf{Question:} \{question\}

\textbf{Retrieved evidence:}

\textbf{[1] \{title\_1\}}\\
\{text\_1\}

\textbf{[2] \{title\_2\}}\\
\{text\_2\}

\hspace*{1em}$\cdots$
\end{nlpgprompt}

The application prompts are unchanged across policy versions. NLPG modifies
only the policy addendum appended to the corresponding module context.

\subsection{Module-Local Policy Injection}
\label{app:nlpg-policy-augmentation}

For module invocation $(i,j)$, NLPG retrieves a bounded set of instructions
from the current policy version and appends them to the original module
context. Let $c_{i,j}^{\mathrm{base}}$ be the original context and let
$\mathcal{P}_{t,i,j}$ be the selected policy instructions. The resulting
context is

\begin{equation}
c_{i,j}
=
c_{i,j}^{\mathrm{base}}
\mathbin{\Vert}
\operatorname{Format}
\left(
\mathcal{P}_{t,i,j}
\right),
\label{eq:prompt-augmentation}
\end{equation}

where $\Vert$ denotes text concatenation.

\begin{nlpgprompt}{Module-Local NLPG Guidance}
\textbf{NLPG procedural guidance for the \{module\_name\} module:}

Use the following instructions to improve the current module operation while
preserving the fixed input and output interface.

\textbf{1.} \{instruction\_1\}\\
\textbf{2.} \{instruction\_2\}\\
\hspace*{1em}$\cdots$
\end{nlpgprompt}

The instructions are ranked by their accumulated cross-execution support and
retained within the configured policy budget. Route-specific instructions are
associated with the module that produced the relevant execution evidence.
Shared instructions may be included when the execution record supports a
correction that applies across modules.

\subsection{LLM-Based Policy Reflection}
\label{app:llm-policy-reflection}

After an execution is evaluated, the critic LLM analyzes its observable
execution record. The record contains the task input, module dependencies,
module inputs and outputs, retrieved evidence, generated queries, final
response, and task-level evaluation feedback. The critic identifies an
observable deviation, attributes it to a module present in the execution, and
proposes a reusable behavioral correction.

\begin{nlpgprompt}{Natural-Language Policy Reflection}
Analyze the evaluated execution and its recorded module dependencies.

Identify the observable decision that contributed to the failure. Select the
responsible module from the modules present in the execution graph. Explain
the local problem and propose one concise procedural instruction that could
help this module avoid the same failure on future task instances.

The instruction must describe reusable behavior. Do not copy task-specific
answers, names, dates, labels, or reference information into the instruction.

Return a structured object containing:

\begin{itemize}
\item an execution-deviation descriptor;
\item the responsible module;
\item a concise critique;
\item one reusable policy instruction;
\item a confidence score; and
\item references to the execution fields supporting the diagnosis.
\end{itemize}

\textbf{Task-level feedback:} \{evaluation\_feedback\}

\textbf{Execution graph:} \{execution\_graph\}

\textbf{Module records:} \{module\_records\}
\end{nlpgprompt}

For invocation $(i,j)$, the resulting textual policy gradient is represented
as

\begin{equation}
g_{i,j}
=
\left(
a_{i,j},
s_{i,j},
u_{i,j},
\kappa_{i,j}
\right),
\label{eq:app-policy-gradient}
\end{equation}

where $a_{i,j}$ is the receiving module selected from the observed execution
graph, $s_{i,j}$ is the deviation descriptor generated by the critic,
$u_{i,j}$ is the proposed natural-language correction, and
$\kappa_{i,j}$ is the critic confidence. The descriptor and correction are
generated from the execution evidence; they are not retrieved from a
benchmark-specific correction table.

\subsection{Execution-Graph Diagnosis and Route Attribution}
\label{app:failure-signatures-routes}

NLPG performs attribution over the module-dependency graph recorded during
each evaluated execution. The evaluator provides task-level feedback, while
the execution record provides the observable structure needed to determine
where that feedback should be propagated.

The backward pass starts at the terminal module whose output contributes
directly to the evaluated response. For an intermediate module, the critic
collects the feedback messages returned by its immediate successors and
examines them together with the module's input, output, and execution
context. The critic may generate a local correction, forward a more specific
feedback message to a recorded predecessor, or stop propagation when the
available evidence does not justify further attribution.

This operation determines the receiving route from the observed execution
graph rather than from a fixed list of benchmark-specific routes. The graph
constrains where feedback may travel, while the critic determines which
observable behavior should change and how the correction should be expressed.
The resulting local corrections are subsequently grouped across task
instances by the policy optimizer.

\begin{nlpgprompt}{Execution-Graph Backward Reflection}
Review the current module and the feedback received from its immediate
successors.

Use the module input, output, and execution context to determine whether the
current module contributed to the downstream error.

Return a local correction only when the execution evidence supports one.
Forward feedback only to modules listed as immediate predecessors in the
execution graph. Do not attribute an error to an unobserved module.

\textbf{Current module:} \{module\_record\}

\textbf{Downstream feedback:} \{downstream\_feedback\}

\textbf{Immediate predecessors:} \{predecessors\}
\end{nlpgprompt}

The backward pass is local to one execution. It should not be confused with
the cross-execution aggregation described in
Section~\ref{sec:nlpg_aggregation}, which combines corrections produced from
different task instances.

\subsection{Policy Initialization and Update}
\label{app:seed-policies}

To separate learned corrections from benchmark-specific behavioral knowledge,
the main NLPG configuration initializes the route-local policy memory without
failure-specific instructions:

\begin{equation}
M_0^a=\varnothing,
\qquad a\in\mathcal{A}.
\label{eq:empty-policy-initialization}
\end{equation}

The fixed application prompts provide only the input and output contracts
required by the program. They do not specify a particular bridge strategy,
query-rewriting rule, retrieval correction, answer-selection heuristic, or
failure-to-policy mapping.

After a group of task instances has been evaluated, the critic generates
textual corrections from the corresponding execution records. The optimizer
then groups compatible corrections, removes instance-specific content,
applies the policy-safety checks, and ranks the remaining candidates by
cross-execution support. The next policy version is

\begin{equation}
M_{t+1}
=
\operatorname{Update}
\left(
M_t,
\{g_{i,j}\}_{i=1}^{N}
\right),
\label{eq:policy-version-update}
\end{equation}

where only corrections supported by the construction data and accepted by the
policy-selection procedure are retained.

A policy instruction therefore enters the memory through the following
sequence:

\begin{equation}
\text{evaluated execution}
\rightarrow
\text{LLM diagnosis}
\rightarrow
\text{module-local correction}
\rightarrow
\text{cross-execution consolidation}
\rightarrow
\text{policy validation}.
\label{eq:policy-construction-sequence}
\end{equation}

The final policy version is frozen before evaluation on the held-out test
partition.

\subsection{Structured Module Outputs}
\label{app:prompt-json}

The application modules return explicit intermediate fields so that later
modules can consume observable outputs and so that NLPG can analyze the
execution trace.

\begin{nlpgprompt}{Module Output Contracts}
\textnormal{Evidence summarization:} summary and salient entities or
relations.

\textnormal{Query generation:} retrieval query and query rationale.

\textnormal{Answer synthesis:} final answer and evidence identifiers.

\textnormal{Execution record:} module inputs, module outputs, route
identifiers, dependencies, and evaluation metadata.
\end{nlpgprompt}

These fields expose the program-level execution record without requiring
access to hidden model reasoning. The policy-update process operates on these
observable fields and does not modify the evaluator, scoring function, or
program control flow.

\subsection{Prompt and Policy Budget}
\label{app:prompt-runtime}

For each module invocation, NLPG retains at most $K$ policy instructions and
limits the rendered policy addendum to the configured character budget. The
same policy budget is used when evaluating a policy version and when
constructing its successor.

The active policy version remains fixed throughout one evaluation pass. After
the pass is complete, the critic analyzes the recorded executions and the
optimizer constructs a candidate successor policy. The candidate is evaluated
on the policy-selection partition before it can be promoted. The promoted
policy is then frozen for the held-out test evaluation.

This separation ensures that a correction generated from one task instance
cannot change the execution of another task instance in the same evaluation
pass, and that final-test executions do not participate in policy construction
or policy selection.

\end{document}